\pdfoutput=1

\documentclass[11pt]{article}

\usepackage[final]{acl}

\usepackage{times}
\usepackage{latexsym}

\usepackage[T1]{fontenc}

\usepackage[utf8]{inputenc}

\usepackage{microtype}
\usepackage{float}

\usepackage{inconsolata}

\usepackage[most]{tcolorbox}
\tcbuselibrary{listings}
\usepackage{graphicx,float} 

\usepackage{inconsolata,paralist}
\usepackage[colorinlistoftodos]{todonotes}

\definecolor{Red}{rgb}{0.7,0.01,0.01}
\definecolor{Purple}{rgb}{.627,.124,.941}

\usepackage{graphicx}
\usepackage{array}
\usepackage{subcaption}
\usepackage{booktabs}
\usepackage[most]{tcolorbox}
\usepackage{multirow}

\tcbuselibrary{listings}
\usepackage{arydshln}
\usepackage{amsfonts}

\newcommand*\rot{\rotatebox{90}}

\title{SemEval-2025 Task 4: \\ Unlearning sensitive content from Large Language Models}
\author{Anil Ramakrishna\textsuperscript{1}, Yixin Wan\textsuperscript{2}, Xiaomeng Jin\textsuperscript{3}, Kai-Wei Chang\textsuperscript{1,2}, Zhiqi Bu\textsuperscript{1}, \\ \textbf{Bhanukiran Vinzamuri}\textsuperscript{1}, \textbf{Volkan Cevher}\textsuperscript{1,4}, \textbf{Mingyi Hong}\textsuperscript{1,5}, \textbf{Rahul Gupta}\textsuperscript{1} \\
\textsuperscript{1}Amazon AGI, \textsuperscript{2}UCLA, \textsuperscript{3}UIUC, \textsuperscript{4}EPFL, \textsuperscript{5}University of Minnesota \\
}

\begin{document}

\maketitle

\begin{abstract}
    We introduce SemEval-2025 Task 4: unlearning sensitive content from Large Language Models (LLMs). The task features 3 subtasks for LLM unlearning spanning different use cases: (1) unlearn long form synthetic creative documents spanning different genres; (2) unlearn short form synthetic biographies containing personally identifiable information (PII), including fake names, phone number, SSN, email and home addresses, and (3) unlearn real documents sampled from the target model’s training dataset. We received over 100 submissions from over 30 institutions and we summarize the key techniques and lessons in this paper. 
\end{abstract}

\section{Introduction}

Large Language Models (LLMs) have achieved enormous success recently due to their ability to understand and solve various non-trivial tasks in natural language. However, they have been shown to memorize their training data \cite{carlini2019secret} which, among other concerns, increases risk of the model regurgitating creative or private content. Often, such issues are discovered post model training during testing or red teaming. Furthermore, stakeholders may sometimes request to remove their data after model training to protect copyright, or exercise their right to be forgotten \cite{GDPR}. In these instances, retraining models after discarding such data is one option but doing so after each such removal request is prohibitively expensive. 

Machine unlearning is a promising line of research for removing sensitive information from models’ parametric memory. 
While unlearning has been studied for sometime in classification problems, it is still a relatively underdeveloped area of study in LLM research since the latter operate in a potentially unbounded output label space. 
Current algorithms often fall short of effectively and efficiently unlearning sensitive information from LLMs, without impacting model utility. Further, there is a need for benchmarks which can provide thorough evaluations of new unlearning algorithms in removing different categories of sensitive information. 

To address these needs and to spur further research on this topic, we developed a new challenge (and an associated benchmark) for LLM Unlearning as part of the SemEval 2025 competition. This document provides a summary of our challenge\footnote{llmunlearningsemeval2025.github.io} along with the benchmark, results and key takeaways. 



\section{Challenge Description}
To robustly evaluate unlearning algorithms on their effectiveness in removing different categories of information from LLM, we developed\footnote{github.com/amazon-science/lume-llm-unlearning} a new unlearning benchmark covering three distinct sub-tasks spanning (1) creative content, (2) Personally Identifiable Information (PII) of synthetic individuals and (3) real biographies of individuals sampled from Wikipedia. Please refer to \cite{ramakrishna2025lumellmunlearningmultitask} for detailed information on the dataset creation process. 

Within each sub-task, we further test the models for regurgitation (where model is prompted to complete partial documents) and knowledge (via generated question-answer pairs), leading to 12 different sub-tasks for the challenge.  To score highly in the challenge, participants are expected to do well in all sub-tasks.
In comparison, existing unlearning benchmarks such as TOFU~\cite{maini2024tofu} and MUSE~\cite{shi2024musemachineunlearningsixway} cover only a portion of the subtasks we test for. 

For each subtask, we released \textit{Retain} ($R$) (i.e. model should retain these documents in memory) and \textit{Forget} ($F$) datasets (i.e. model should forget information from these documents) along with two target models (7 billion and 1 billion parameters in size) which were fine-tuned to memorize documents from all three tasks. 

Participants were encouraged to explore various unlearning algorithms which enable them to remove the sensitive information present in $F$ without affecting model knowledge on the $R$. Our initial data release was further split in 80:20 ratio as train and validation subsets for optional hyper-parameter tuning. 
Participants were expected to submit working Python scripts containing their unlearning code for the evaluation phase, which were executed on privately held subsets of retain and forget sets from each sub-task. 
Table \ref{tab:statistics} lists overall statistics of our benchmark, and examples are shown in Figure \ref{fig:lume_approach}.

We provide further details on our dataset creation for the three tasks below, followed by details on the evaluation phase. 

\subsection{Dataset Creation}

\begin{table}
\centering
\begin{tabular}{c:cc:c}
 & Forget & Retain &    \\ \midrule
Task 1 & 199    & 194    & 393 \\
Task 2 & 203    & 202    & 405 \\
Task 3 & 295    & 294    & 589 \\ \hdashline
       & 697    & 690    & 1,387  \\ 
\end{tabular}
\caption{Number of unique documents for both data subsets within each task. For each document, we create multiple regurgitation and knowledge datasets leading to 4,394 unique examples. }
\label{tab:statistics}
\end{table}

\subsubsection{Task 1: Synthetic creative documents}
LLMs trained on Internet-scraped data are often exposed to copyrighted content, making unlearning of this information a common requirement post training. However, evaluating effectiveness of unlearning on only real creative documents \cite{shi2024musemachineunlearningsixway,eldan2023whosharrypotterapproximate} is challenging as information to be removed may appear in other documents not being unlearned. For example, MUSE \cite{shi2024musemachineunlearningsixway} uses Harry Potter books as its forget set, but this information may be exposed to the model via Wikipedia articles and social media posts. Motivated by this, in this task, we only include synthetically generated short novels, created using Mixtral 8x7B \cite{jiang2023mistral} as our generator LLM.

To create each document in this task, we first randomly sample a genre from one of Action, Fantasy, Thriller, Comedy, Mystery, Science Fiction, Young Adult and Romance. Next we generate one to four synthetic character names using a random name generator \footnote{pypi.org/project/unique-names-generator}, and synthetic locations from the city list of a random address generator \footnote{pypi.org/project/random-address} for all genres except Fantasy genre. For Fantasy, we sample unique genre specific city names using a Dungeons and Dragons town generator \footnote{perchance.org/dndtowngen}. Given this information, we prompt the Mixtral model (full prompt listed in Appendix \ref{sec:long_form_section}) to create a short story with 150-200 words. To validate the generated stories, we conducted manual reviews (each short story was reviewed by two authors) and filtered out stories with similar content to prior reviewed stories. Our final dataset for this task contained 393 unique short stories across all genres.

\subsubsection{Task 2: Synthetic biographies with sensitive PII}
We use various heuristics to generate 500 synthetic personal biographies with following PII fields: 

\begin{compactitem}
    \item \textit{Name}: randomly created from a name generator, includes firstname+lastname.
    \item \textit{Birthday}: randomly sampled between 01/01/1964 and 01/01/1991.
    \item \textit{Social Security number (SSN)}: randomly sampled within the range 900-xx-xxxx (which by policy cannot not belong to a real person \cite{SocialSecurityChanging}). 
    \item \textit{Phone Number}: 10 randomly sampled digits.
    \item \textit{Email address}: Created heuristically of the form \texttt{firstname\_lastname@me.com}. 
    \item \textit{Home address}: A non-existent physical home addresses obtained by combining a random street address from a US state with an alternate city and zip-code from a different state. 
\end{compactitem}

For each synthetic individual created above, we prompt the Mixtral model (using prompt listed in Appendix \ref{sec:short_form_section}) to create a short biography which includes all the PII information.

\subsubsection{Task 3: Real biographies}
To evaluate effectiveness of unlearning on real data, we include real biographies as the third task. Specifically, we sampled 750 biographies spanning 100 to 200 words from Wikipedia documents released in the Dolma \cite{dolma} v1.6 corpus, which was part of the training dataset for the OLMo models~\cite{Groeneveld2023OLMo} we use for this task.

\subsection{Subtasks}
For each task, we additionally created prompts for two subtasks detailed below. 

\subsubsection{Regurgitation tests}
To test for model regurgitation of documents, we created sentence completion prompts for all documents from the three  tasks by sampling a random position in second half of the document with the sentences before it as the input.

\subsubsection{Knowledge tests}
We create question answer prompts for each document using an agentic workflow for Tasks 1 and 3 where we prompt the data generator LLM (Mixtral 8x7b) with few-shot Chain of Thought prompting \cite{wei2022chain} (prompt listed in Appendix \ref{sec:question_generation_prompt}) to construct an unambiguous question with a single concise answer. We validate the quality of the generated QA pair by prompting three verification LLMs (Claude 3 Sonnet, Titan Text Express and Mixtral 8x7B) to answer the question with full document as grounding. We discard QA pairs if any of the three verification LLMs are unable to answer the question accurately. For Task 2, we use template based heuristics for each PII field to frame questions of the form \textit{What is the birth date of John Smith?} with the corresponding entry as the answer. 

\subsection{Data Splits}

We divide the dataset we created into two halves, corresponding to forget ($F$) and retain ($R$) subsets. Each unlearning algorithm is evaluated on how well it can erase sensitive information from the forget subset, without impacting information in the retain subset. We maintain a 1:1 ratio between the two subsets, which adds to the challenge. We further split both of these into private and public subsets. We released the public retain and forget subsets in September 2024, as part of the task artifacts. The private datasets were saved for the evaluation phase. 

\subsection{Unlearning Model Candidates}

We fine-tuned OLMo-7B-0724-Instruct-hf (7 Billion parameters\footnote{huggingface.co/llmunlearningsemeval2025organiza\\tion/olmo-1B-model-semeval25-unlearning}) and  OLMo-1B-0724-hf (1 Billion parameters\footnote{huggingface.co/llmunlearningsemeval2025organiza\\tion/olmo-finetuned-semeval25-unlearning}) models on documents from all three tasks and release them as unlearning candidates. We selected OLMo because of its permissive license and open sourced training dataset (with logs) which enables downstream task specific analyses of model behavior. 

\subsection{Evaluation}

In typical evaluation cycles, participants are invited to upload their trained model checkpoints which are evaluated on a private test set. However, since unlearning algorithms need access to the targeted information to erase from the model’s memory, we would have to release the private forget and retain subsets. But this can compromise the integrity of evaluations since a participant may chose to retrain the OLMo models from scratch on just the retain data subsets, achieving high scores in our evaluation metrics. 

To avoid this, in our challenge we invited each participant to develop their unlearning algorithms locally using the publicly released forget and retain subsets and upload their working code for evaluation. For each such submission, we individually call the corresponding unlearning functions with the private forget and retain subsets as arguments, and evaluate the generated checkpoints for unlearning effectiveness. During the evaluation phase, submissions were timed and those runs which take more than a pre-determined threshold of time were discarded. Further, to support diverse explorations, each team was invited to submit up to 5 distinct code files for the evaluation, of which the best performing candidate (among those which finished training and retained model utility) was selected for the leaderboard. 

All evaluation experiments were conducted (with limited permissions) on an AWS EC2 p4d.24xlarge node with 8 A100 40 GB GPUs. The compute environment was pre-configured with DeepSpeed Zero \cite{rajbhandari2020zeromemoryoptimizationstraining} with additional packages installed if requested by the teams. 

To evaluate the generated checkpoints, we computed following metrics:

\subsubsection{Task memorization metrics}

For each of our three tasks, we compute two distinct metrics listed below, corresponding to the two subtasks to evaluate the model’s memorization of sensitive information: 

\noindent \textbf{a) Regurgitation Rate}: We compute ROUGE-L  \cite{lin-2004-rouge} scores for the model generated outputs with respect to the expected sentence completions. We chose rouge since it is weighted for recall of sensitive information in model outputs. 

\noindent \textbf{b) Knowledge Test Accuracy}: For all QA prompts, we use case insensitive exact match between model output and the expected answer to measure prediction accuracy.

Overall, we compute 12 different metrics which measure memorization.  We compute the harmonic mean of these to obtain a single task-aggregate metric. 

\begin{figure*}
    \centering
    \begin{subfigure}[b]{0.22\textwidth}{\includegraphics[scale=0.24]{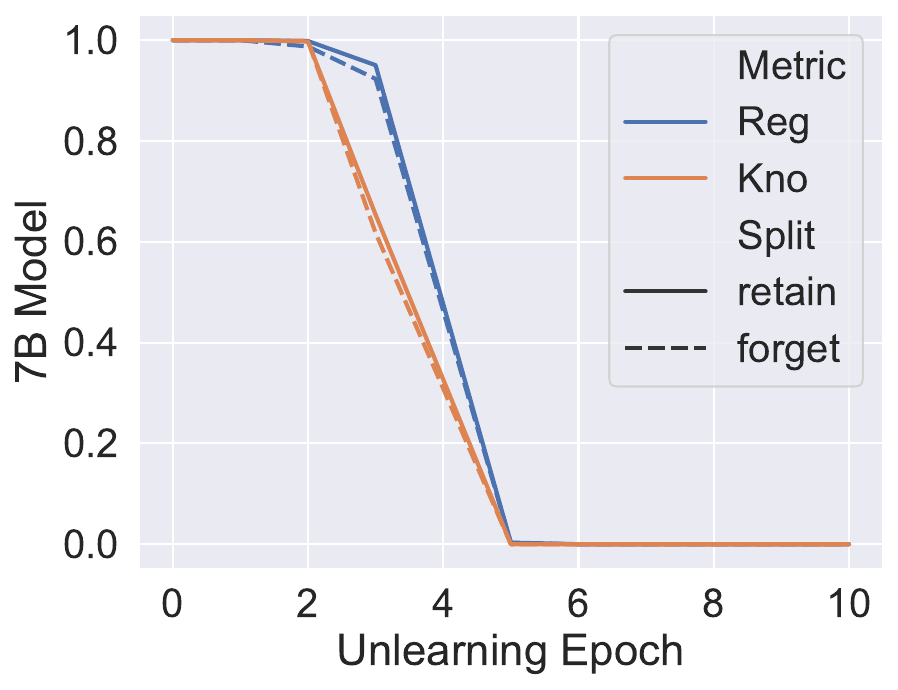}}\end{subfigure}
    \begin{subfigure}[b]{0.22\textwidth}{\includegraphics[scale=0.24]{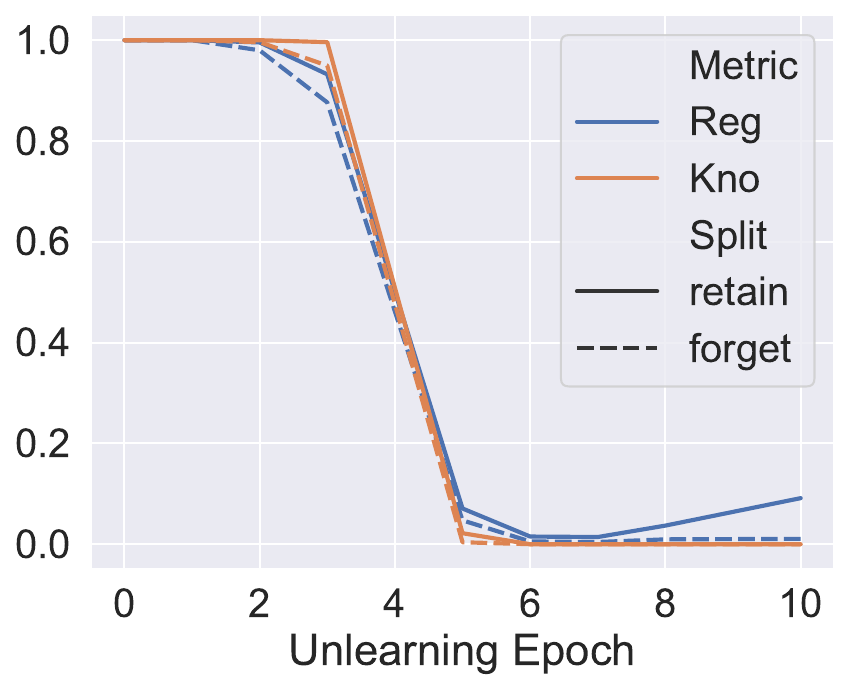}}\end{subfigure}
     \begin{subfigure}[b]{0.22\textwidth}{\includegraphics[scale=0.24]{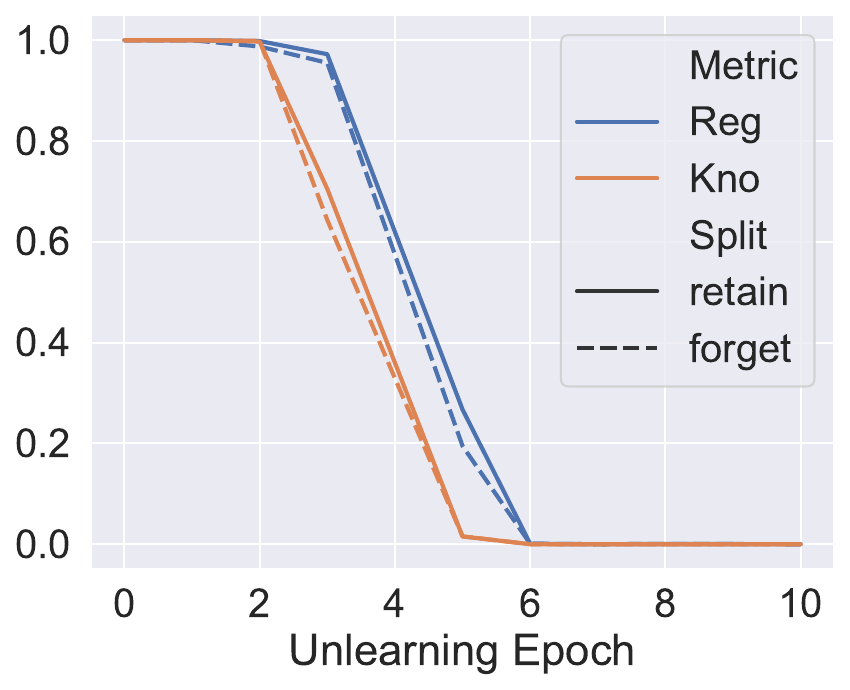}}\end{subfigure}
     \begin{subfigure}[b]{0.22\textwidth}{\includegraphics[scale=0.24]{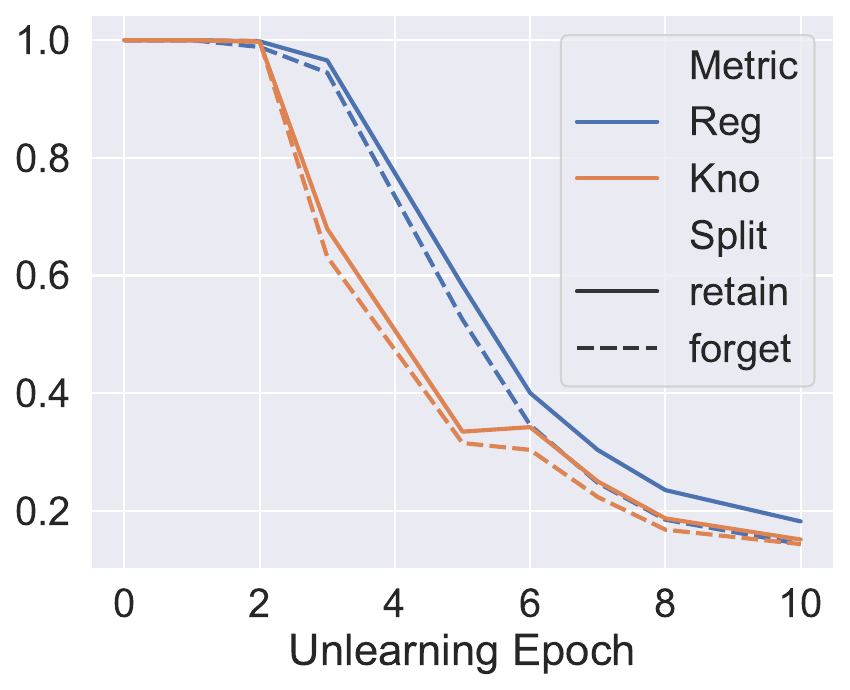}}\end{subfigure}
    \begin{subfigure}[b]{0.22\textwidth}{\includegraphics[scale=0.24]{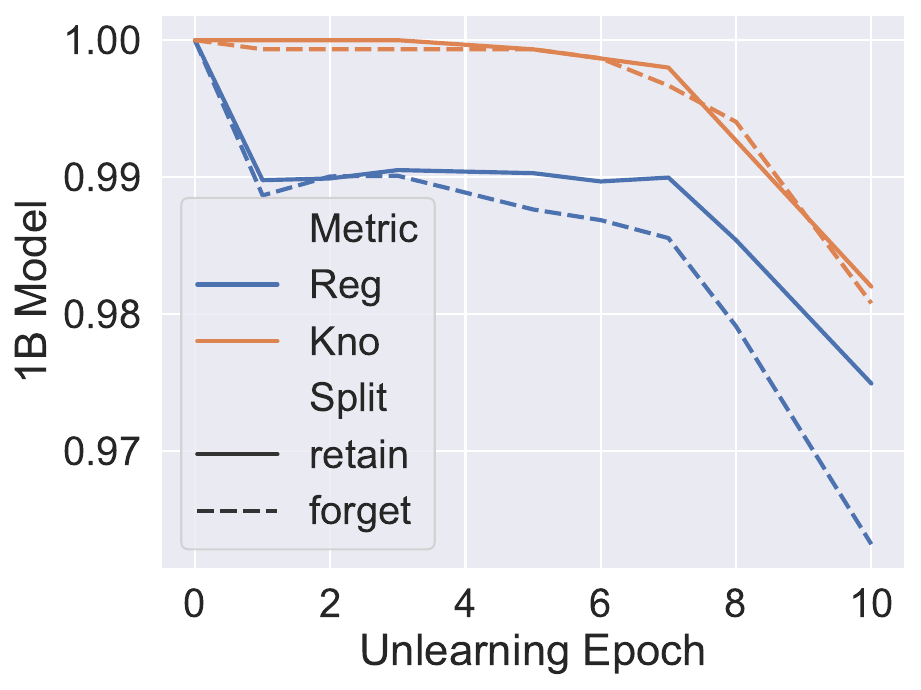}}\caption{GA}\end{subfigure}
    \begin{subfigure}[b]{0.22\textwidth}{\includegraphics[scale=0.24]{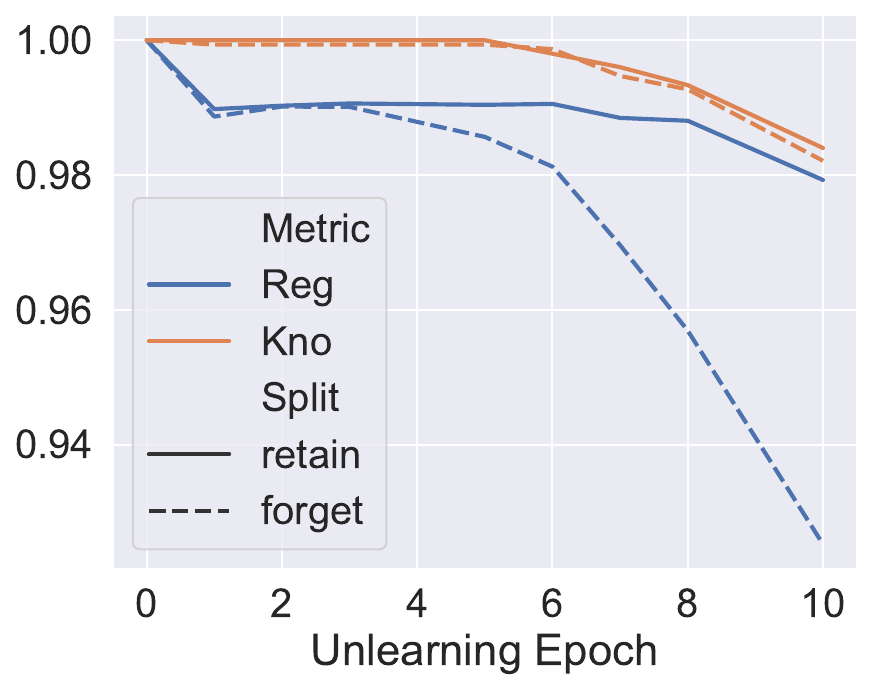}}\caption{GD}\end{subfigure}
     \begin{subfigure}[b]{0.22\textwidth}{\includegraphics[scale=0.24]{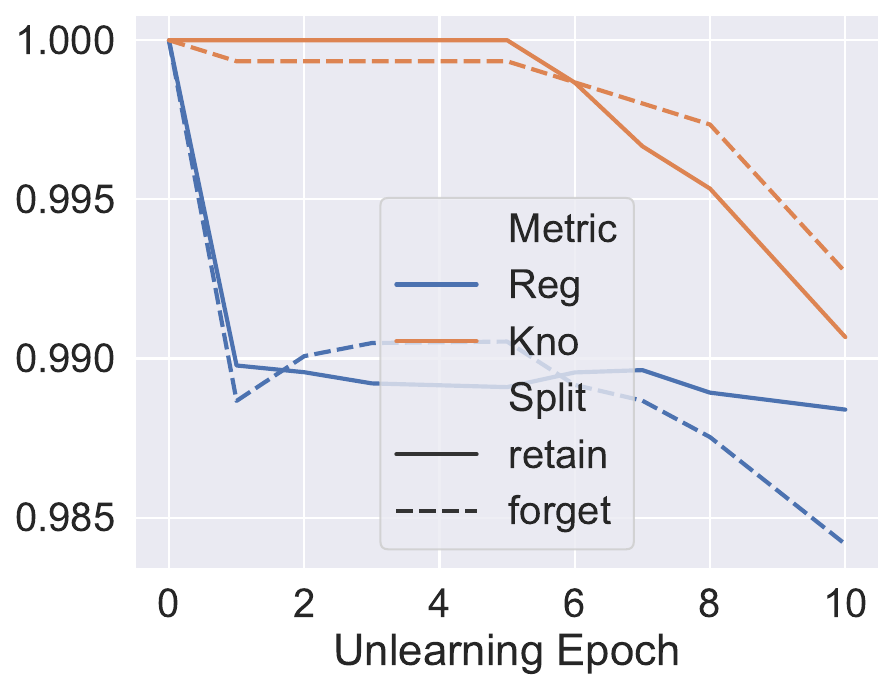}}\caption{KL}\end{subfigure}
     \begin{subfigure}[b]{0.22\textwidth}{\includegraphics[scale=0.24]{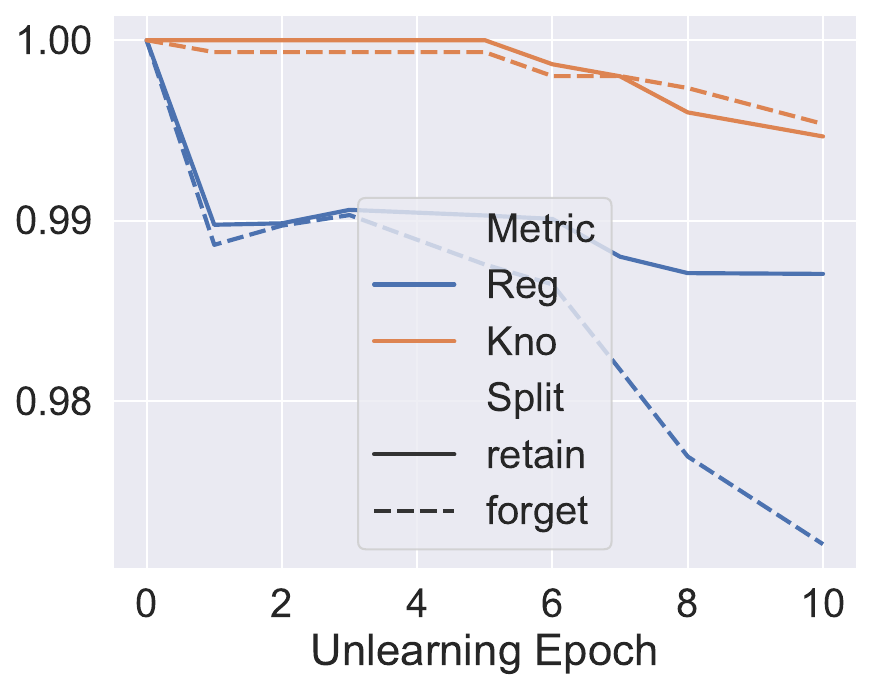}}\caption{NPO}\end{subfigure}
    \caption{Performance on \textit{retain} and \textit{forget} subsets for benchmarked unlearning algorithms. Reg: Regurgitation Rate ($r$), Kno: Knowledge Accuracy ($t$). Split refers to data subset (forget or retain) used in evaluations.}
    \label{fig:benchmark_results}
\end{figure*}

\subsubsection{Membership Inference Attack success rates (MIA)}

Since the model may retain some sensitive information despite showing low memorization rates after unlearning, we also compute MIA rates on the subtask prompts. We compute loss based membership inference attacks using the MIA attack framework from ~\cite{duan2024membership} to assess data leakage risk after unlearning. A robust unlearning algorithm should effectively remove evidence of the forget set and yield MIA success rates close to $0.5$ AUC (random chance) between member v/s non-member datasets. We use a subset of the memorized Wikipedia biographies from the forget subset of Task 3 as the member set and a disjoint sample of similar biographies not exposed to the model as the non-member set. Further, we compute following MIA score to penalize any deviations from $0.5$:
\begin{equation*}
    \text{MIA Score} = 1 - 2 \cdot \left| \text{mia\_auc} - 0.5 \right|
\end{equation*}
\subsubsection{Model Utility}

We also test for overall model utility by computing test set accuracy for 57 STEM subjects from MMLU \cite{hendryckstest2021}, a general benchmark for LLM utility. We also threshold on this metric for the post unlearning candidate to avoid trivial solutions which completely distort general model utility but achieve high scores in the task aggregate and/or MIA (such as Gradient Ascent). Specifically for the 7B model, we only consider submissions for which the MMLU accuracy is above 0.371 (75\% of the pre-unlearning checkpoint) for our official awards leaderboard. However, we did not impose this constraint for the 1B model since the performance of the base model on this dataset was already low, close to random  chance. 

\subsubsection{Aggregate Final Score}

Finally, we compute arithmetic mean of the task aggregate metric, MIA score and the model utility to obtain a single numeric score to compare all submissions.

\section{Benchmarked Algorithms}
\label{sec:unlearning-method}

We benchmarked our dataset on several state of the art unlearning algorithms described below. 

\noindent \textbf{Gradient Ascent}: This is a straightforward unlearning algorithm where we reverse the direction of model update by flipping the sign in gradient descent, in order to steer the model away from the sensitive model outputs in the forget set. While easy to implement, this approach has a significant drawback since the gradient ascent training objective is unbounded, which can lead to model divergence with nonsensical outputs for all inputs. The loss term in this algorithm reverses sign of the standard Cross Entropy training loss ($\mathcal{L}_{CE}$) and is applied only on the forget set $F$: 
\begin{align*}
     -\mathcal{L}_{CE}(F; \theta)
\end{align*}

\noindent \textbf{Gradient Difference} \cite{liu2022continual}: In this approach, we augment the gradient ascent objective applied on forget set, by adding a gradient descent objective on the retain set. By jointly optimizing on both sets, we steer the model away from regurgitating the sensitive information from the retain set, while ensuring it does not lose performance in the retain set. Despite being a promising alternative to Gradient Ascent, this quality of model performance on non-sensitive dataset depends on the size of the retain set used in model training, and can lead to poor generalization on new examples. The loss term jointly increases the likelihood of generating responses in the retain set $R$ while reducing the likelihood of generating $F$, as shown below.
\begin{align*}
     -\mathcal{L}_{CE}(F; \theta) + \mathcal{L}_{CE}(R; \theta)
\end{align*}

\noindent \textbf{KL Regularization} \cite{maini2024tofu} Similar to Gradient Difference, in this baseline, we augment the gradient ascent objective with a Kullback-Leibler Divergence term to ensure the model does not deviate too far from the original model. 
\begin{align*}
     -\mathcal{L}_{CE}(F; \theta) + \mathcal{L}_{KL}(R; \theta, \theta_{ref})
\end{align*}

\noindent \textbf{Negative Preference Optimization} \cite{zhang2024negativepreferenceoptimizationcatastrophic}: This baseline uses a modified version of the Direct Preference Optimization objective, adapted to remove the sensitive information from  forget set.
\begin{align*}
     \mathcal{L}_{NPO}(F; \theta)
\end{align*}
\subsection{Benchmark Results}
Consistent with other recent benchmarks, we evaluate each algorithm described above using following hyper-parameters and provide these results to the participants for reference. We use a batch size of 32, and run the algorithms for 10 epochs using a learning rate = $1e-5$ on both models. Figure \ref{fig:benchmark_results} plots their performance on forget and retain sets (task wise plots are shown in Appendix \ref{sec:task_wise_benchmark_results}). We observe over-unlearning with the 7B model but under-unlearning with the 1B model for selected hyper-parameters, suggesting room for improvements by participants over these baselines.


\begin{table*}[]
\centering
\begin{tabular}{l|cccccccccc} 
Team                            & \rot{Gradient Ascent} & \rot{Gradient Difference} & \rot{KL Regularization} & \rot{Other Objectives} & \rot{Random Labels} & \rot{Targeted Unlearning} & \rot{Novel Data Mixing} & \rot{PEFT (LoRA)} & \rot{Model Merging} & \rot{Stabilized Training} \\ \toprule
AILS-NTUA                       &                 & \checkmark                    &                   &     &                              & \checkmark                          &  \checkmark                 &       \checkmark      &               & \checkmark                              \\
ZJUKLAB                         &                 &    \checkmark                 &         \checkmark          &  \checkmark   &                              &                           &                   &       \checkmark      &      \checkmark           &                               \\
YNU            &                 &         \checkmark              &                   &     &        \checkmark                        &                           &                   &             &               &                      \checkmark           \\
Mr. Snuffleupagus               &                 &                     &                   &  \checkmark     &                              &                       \checkmark      &                   &             &               &                               \\
ishumei-Chinchunmei             &                 &                     &                   &   \checkmark    &                              &                           &             \checkmark        &             &               &                               \\
GUIR             &        \checkmark           &                     &       \checkmark              &     \checkmark   &                              &                           &                   &             &               &                               \\
GIL-IIMAS UNAM                  &                 &                     &                   &   \checkmark    &                              &                           &                   &             &               &                               \\
Atyaephyra                      &                 &                     &      \checkmark               &     \checkmark   &                              &                           &                   &           \checkmark    &               &                               \\
Lacuna Inc.                     &                 &                     &                   &     &                              &              \checkmark               &                   &             &               &   \checkmark                              \\
NLPART                          &                 &                       &                   &     \checkmark  
 &                              &                           &                   &             &               &                                \\
JU-CSE-NLP'25                   &                 &    \checkmark                   &                   &     &                              &                           &                   &             &               &                       \checkmark          \\
\bottomrule
\end{tabular}
\caption{Key ideas explored in participating teams, sorted based on their performance on 7B model.}
\label{tab:key_concepts}
\end{table*}

\section{Participant Systems}
We received over 100 submissions from 24 teams with nearly 70 individuals spanning over 30 institutions across the world. We list key ideas explored by participants in Table \ref{tab:key_concepts}. 

Most teams used variations of Gradient Difference (GD), KL Regularization or Negative Preference Optimization (NPO) with specific hyper-parameters coupled with clever optimizations leading to faster training within the fixed compute time. Other teams explored new and innovative solutions for unlearning by leveraging novel loss objectives, selective layer/parameter training, etc. 

The best performing team, \textbf{AILS-NTUA}, leveraged a parameter-efficient unlearning method based on GD with LoRA adapters added to transformer projection layers. They carefully sampled chunks of forget set mixed with a large (resampled) retain set. The second place team, \textbf{ZJUKLAB} merged two different models unlearned with distinct hyperparameters, to balance under/over-unlearning in the two models. The third place team, \textbf{YNU} used alternating GD with randomly sampled forget labels. Team \textbf{Mr. Snuffleupagus} applied targeted unlearning using RMU on 3 layers selected using the validation set. \textbf{ishumei-Chinchunmei} explored a new inverted loss function for the forget set, which avoids the gradient explosion commonly found in GA. 

\textbf{SHA256} use causal mediation analysis on the OLMo models and identify the first five model layers as most relevant for unlearning, and apply re-weighted GD. While this approach achieved high unlearning performance, it considerably degraded model utility on MMLU. 
Team \textbf{Atyaephyra} use LoRA adapters with NPO, regularized using KL, with low memory footmark by offloading the adapters during distillation. However, their submission included an early exit bug during 7B evaluations which led to low performance with this model. This was corrected and resubmitted in time for 1B evaluation, in which their submission took the third spot. 
We present more detailed summaries of the core strategies used by participating teams in Table \ref{tab:strategy}.

\begin{table*}[h!]
\centering
\begin{tabular}{|c|c|c|c|c|}
\hline
\textbf{Team} & \textbf{Final Score} & \textbf{Task Aggregate} & \textbf{MIA Score} & \textbf{MMLU Avg.} \\
\hline
\multicolumn{5}{|c|}{\textbf{Results from 7B Models}} \\
\hline
AILS-NTUA & 0.706 & 0.827 & 0.847 & 0.443 \\
ZJUKLAB & 0.487 & 0.944 & 0.048 & 0.471 \\
YNU & 0.47 & 0.834 & 0.139 & 0.436 \\
Mr. Snuffleupagus & 0.376 & 0.387 & 0.256 & 0.485 \\
ishumei-Chinchunmei & 0.326 & 0.496 & 0 & 0.481 \\
\hline
\multicolumn{5}{|c|}{\textbf{Results from 1B Models}} \\
\hline
AILS-NTUA & 0.688 & 0.964 & 0.857 & 0.242 \\
SHA256 & 0.652 & 0.973 & 0.741 & 0.243 \\
Atyaephyra & 0.586 & 0.887 & 0.622 & 0.248 \\
Mr. Snuffleupagus & 0.485 & 0.412 & 0.793 & 0.25 \\
ZJUKLAB & 0.483 & 0.915 & 0.292 & 0.243 \\
\hline
\end{tabular}
\caption{Scores from the top-5 teams for 7B and 1B models. Complete results are published at \url{llmunlearningsemeval2025.github.io}.}
\label{tab:top-systems}
\end{table*}

\begin{table*}[h!]
\centering
\begin{tabular}{|c|c|c|c|c|c|c|}
\hline
\textbf{Team} & \multicolumn{3}{c|}{\textbf{Regurgitation Score}} & \multicolumn{3}{c|}{\textbf{Knowledge Score}} \\
\hline
& \textbf{Task 1} & \textbf{Task 2} & \textbf{Task 3} & \textbf{Task 1} & \textbf{Task 2} & \textbf{Task 3} \\
\hline
\multicolumn{7}{|c|}{\textbf{Forget Set}} \\
\hline
AILS-NTUA & 0.963 & 0.986 & 0.979 & 0.966 & 0.998 & 0.951 \\
ZJUKLAB  & 0.992 & 0.980 & 0.990 & 1.000 & 1.000 & 1.000 \\
YNU & 0.963 & 0.995 & 0.904 & 0.992 & 1.000 & 0.993 \\
Mr. Snuffleupagus & 0.594 & 0.994 & 0.916 & 0.415 & 1.000 & 0.566 \\
ishumei-Chinchunmei  & 0.587 & 0.634 & 0.637 & 0.603 & 0.567 & 0.601 \\
\hline
\multicolumn{7}{|c|}{\textbf{Retain Set}} \\
\hline
AILS-NTUA & 0.493 & 0.995 & 0.556 & 0.758 & 0.990 & 0.844 \\
ZJUKLAB & 0.671 & 0.952 & 0.815 & 0.527 & 0.799 & 0.696 \\
YNU & 0.896 & 0.981 & 0.749 & 0.967 & 0.984 & 0.970 \\
Mr. Snuffleupagus & 0.485 & 0.290 & 0.145 & 0.582 & 0.167 & 0.526 \\
ishumei-Chinchunmei  & 0.502 & 0.392 & 0.428 & 0.330 & 0.470 & 0.452 \\
\hline
\end{tabular}
\caption{Regurgitation and Knowledge Scores for the top-5 teams on 3 sub-tasks in the 7B model. Higher values indicate better performance in all scores.}
\label{tab:subtask}
\end{table*}

\begin{figure*}[h!]
    \centering
    \begin{subfigure}[b]{0.3\textwidth}{\includegraphics[scale=0.3]{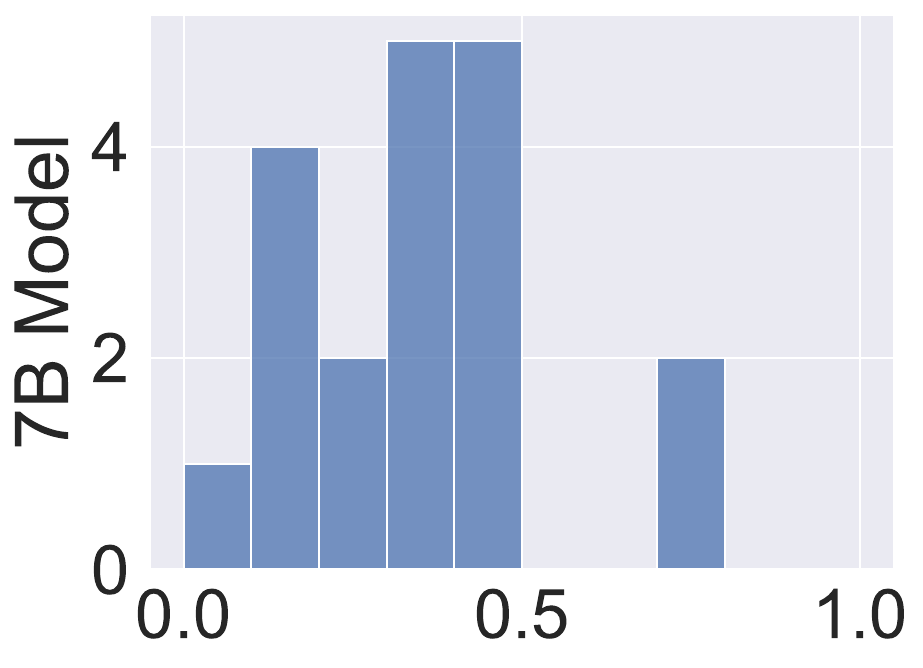}}\label{fig:7B_leaderboard_hist_agg}\end{subfigure}
    \begin{subfigure}[b]{0.3\textwidth}{\includegraphics[scale=0.3]{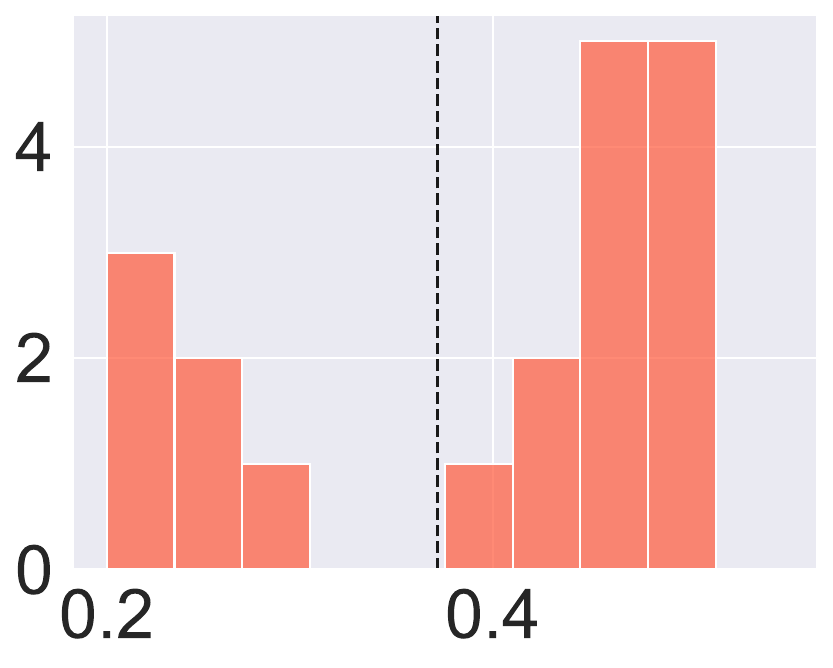}}\label{fig:7B_leaderboard_hist_mmlu}\end{subfigure}
    \begin{subfigure}[b]{0.3\textwidth}{\includegraphics[scale=0.3]{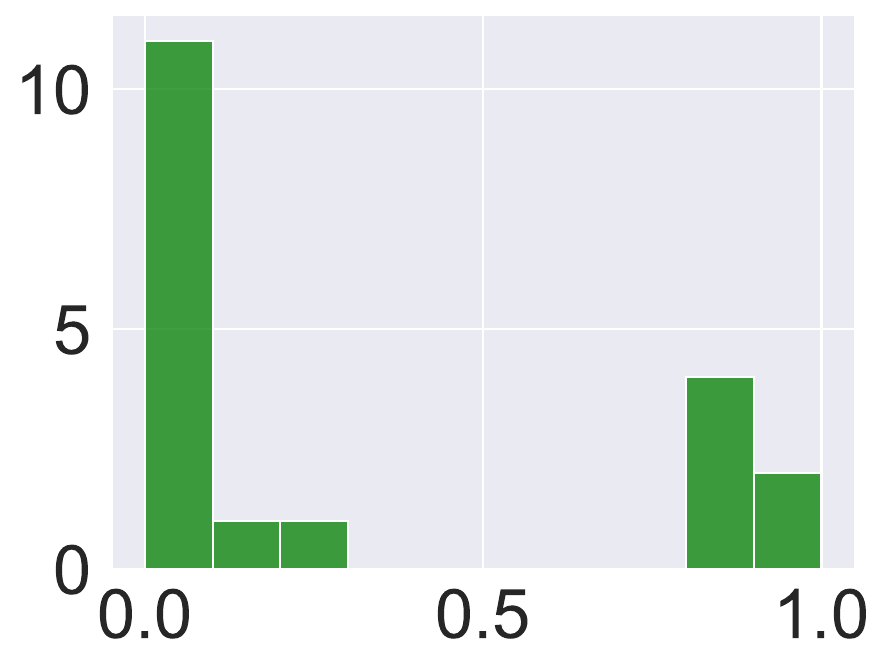}}\label{fig:7B_leaderboard_hist_mia}\end{subfigure}
    
    \centering
    \begin{subfigure}[b]{0.3\textwidth}{\includegraphics[scale=0.3]{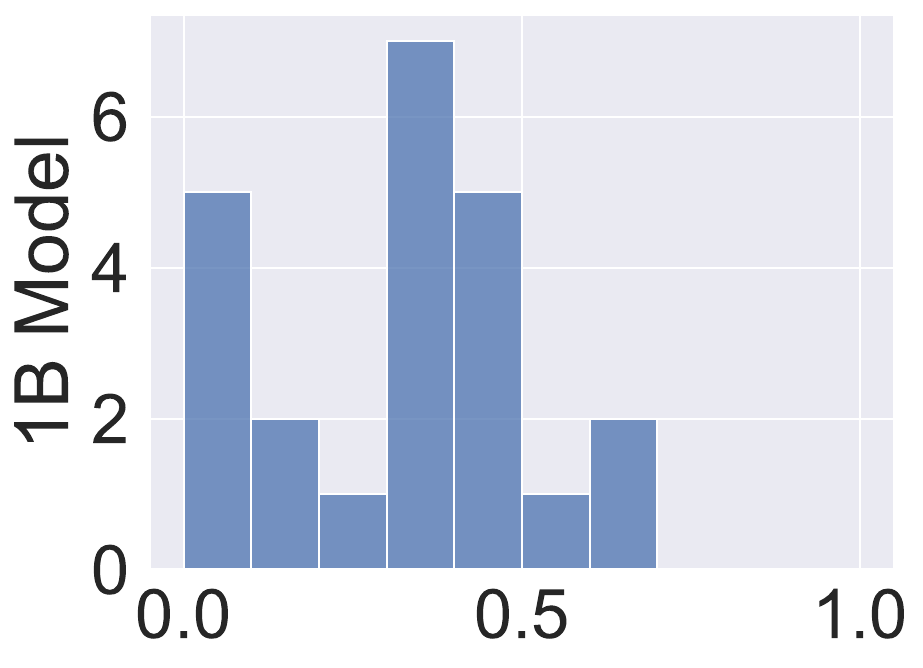}}\label{fig:1B_leaderboard_hist_agg}\caption{Final scores}\end{subfigure}
    \begin{subfigure}[b]{0.3\textwidth}{\includegraphics[scale=0.3]{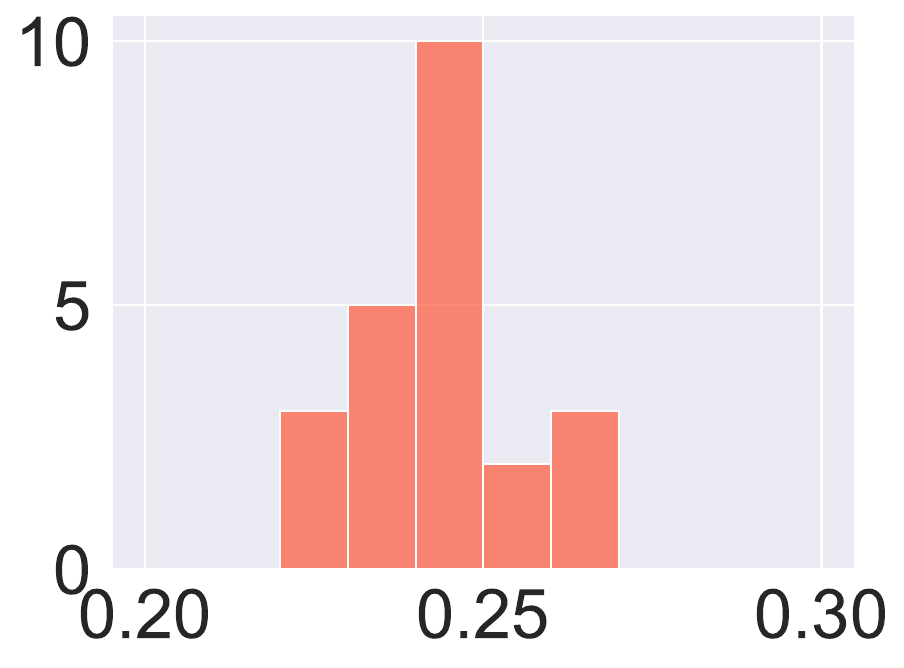}}\label{fig:1B_leaderboard_hist_mmlu}\caption{MMLU}\end{subfigure}
    \begin{subfigure}[b]{0.3\textwidth}{\includegraphics[scale=0.3]{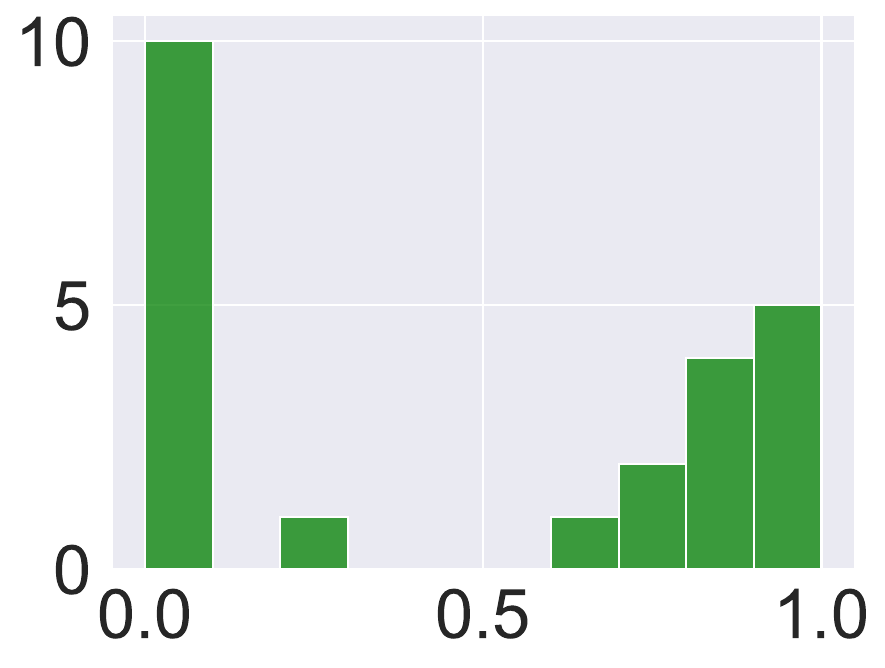}}\label{fig:1B_leaderboard_hist_mia}\caption{MIA}\end{subfigure}
    \caption{Distribution of key scores for all participants on both models. MMLU plots are zoomed in (but still contain 10 bins). Dashed line indicates threshold for 7B model utility below which submissions are discarded.}
 \label{fig:histograms}
\end{figure*}

\subsection{Results and Discussion}

Table \ref{tab:top-systems} presents performance of the top teams when their unlearning algorithms are applied to 7B and 1B models. \textbf{AILS-NTUA} achieved the best performance with both the 1B and 7B models, as their system excels across all three metrics. While \textbf{ZJUKLAB} performed better on Task Aggregate and MMLU scores for the 7B model, their submission significantly underperformed on the MIA score suggesting the unlearned information was not completely removed from model parameter space, and also highlighting a trade-off between MIA and the Task Aggregate scores (also observed in \cite{Ramakrishna2024}). 

Results for both models are largely consistent, with three teams (\textbf{AILS-NTUA}, \textbf{Mr. Snuffleupagus}, and \textbf{ZJUKLAB}) ranking in the top five positions on both leaderboards. As discussed earlier, \textbf{Atyaephyra} had a bug in their submission which was addressed before 1B evaluations thereby gaining several positions. 

\begin{figure*}[h!]
    \centering
    \begin{subfigure}[b]{0.3\textwidth}{\includegraphics[scale=0.3]{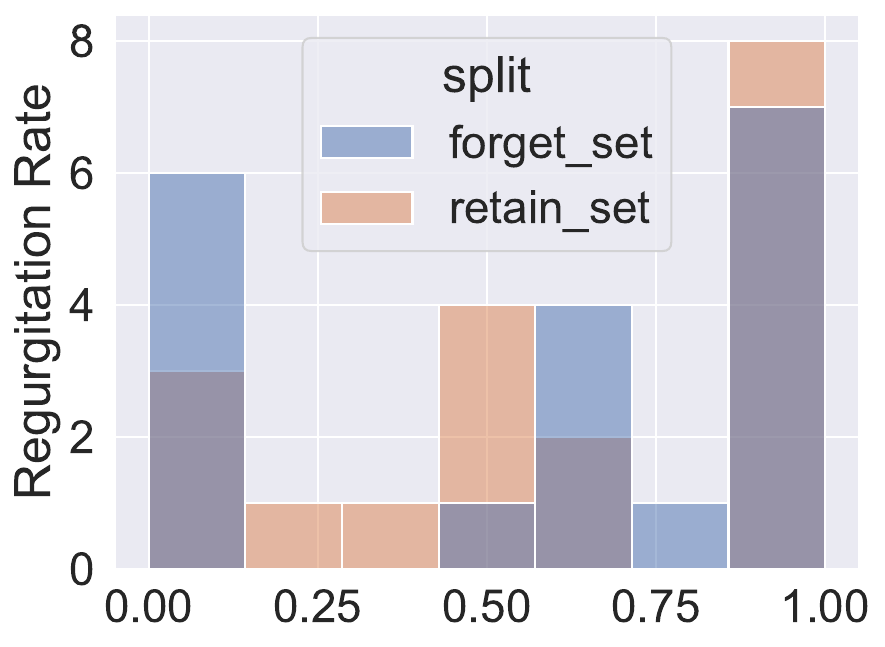}}\end{subfigure}
    \begin{subfigure}[b]{0.3\textwidth}{\includegraphics[scale=0.3]{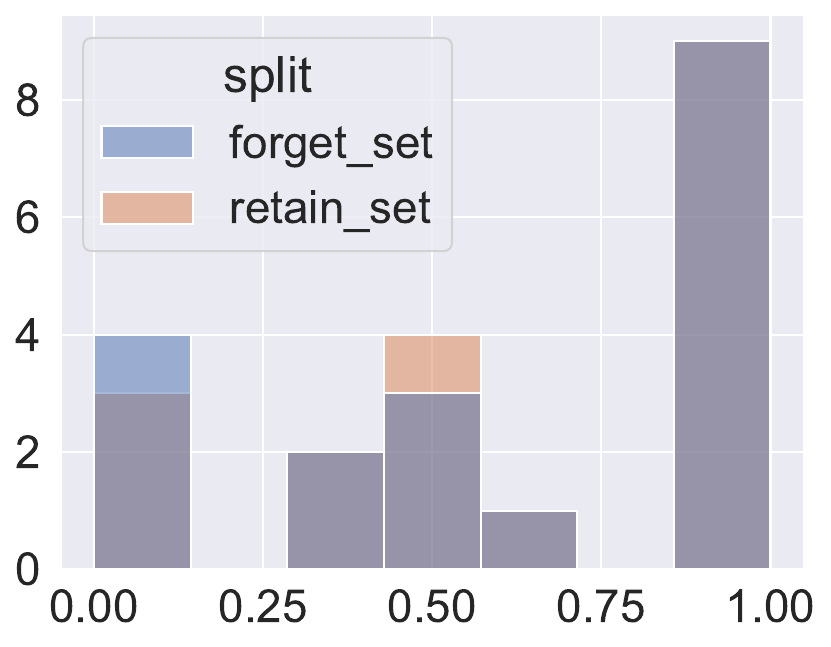}}\end{subfigure}
    \begin{subfigure}[b]{0.3\textwidth}{\includegraphics[scale=0.3]{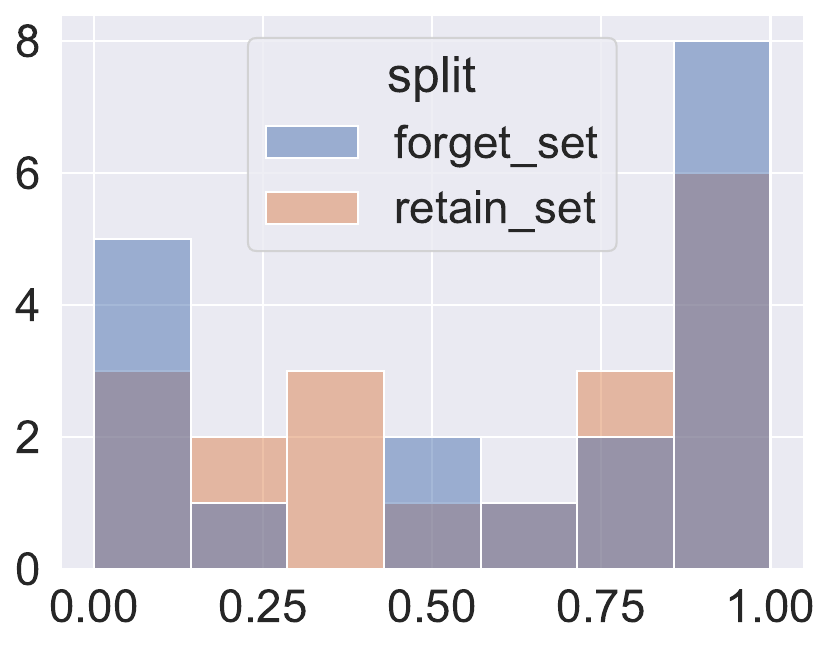}}\end{subfigure}
    
    \centering
    \begin{subfigure}[b]{0.3\textwidth}{\includegraphics[scale=0.3]{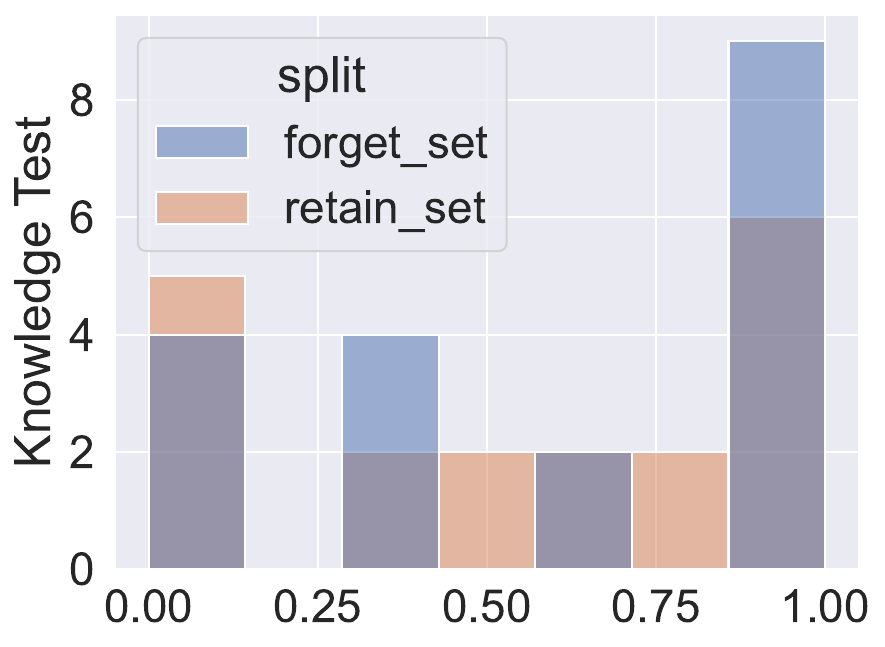}}\caption{Task1}\end{subfigure}
    \begin{subfigure}[b]{0.3\textwidth}{\includegraphics[scale=0.3]{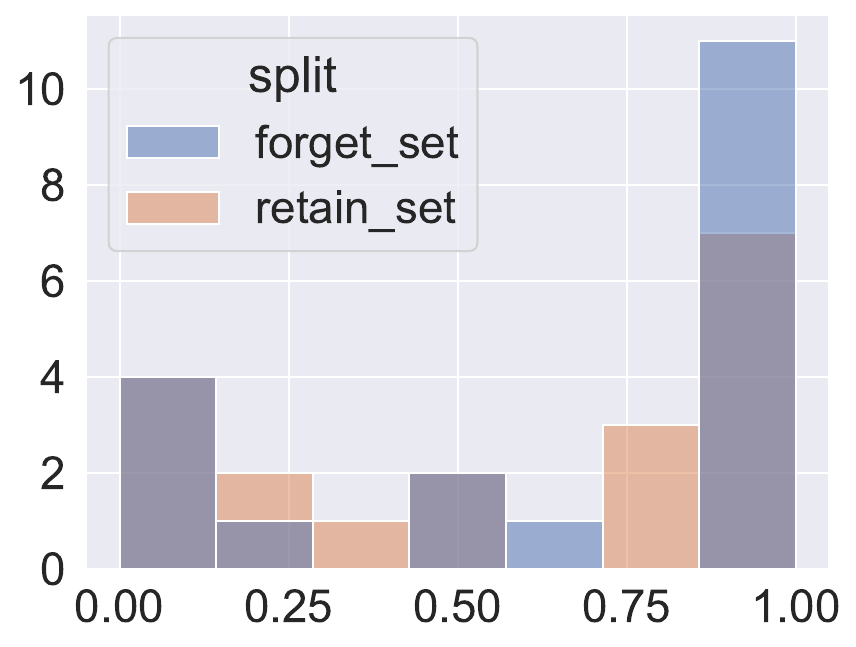}}\caption{Task2}\end{subfigure}
    \begin{subfigure}[b]{0.3\textwidth}{\includegraphics[scale=0.3]{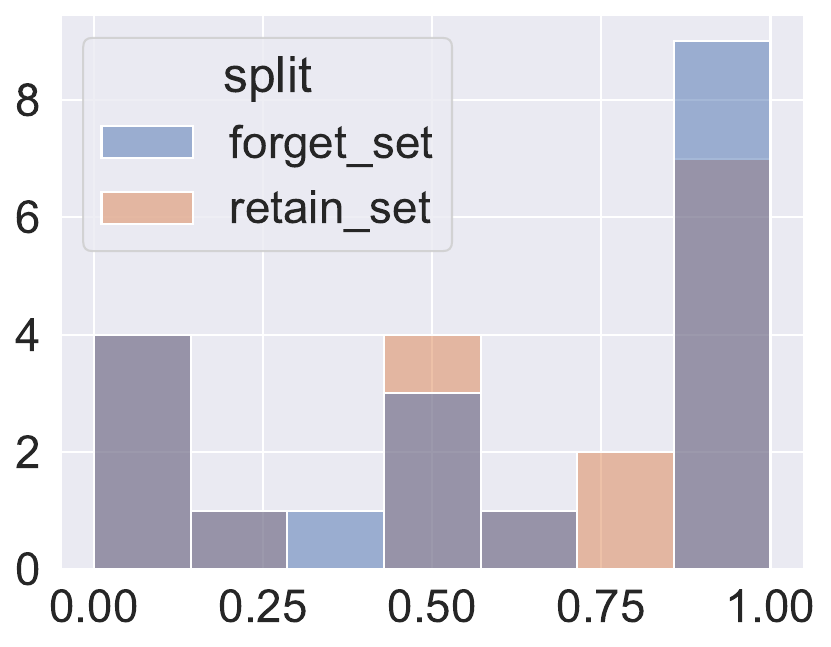}}\caption{Task3}\end{subfigure}
    \caption{Distribution of participant scores for forget and retain sets on the 7B model for all 6 sub-tasks.}
 \label{fig:7B_taskwise_histogram}
\end{figure*}

\begin{figure*}[h!]
    \centering
    \begin{subfigure}[b]{0.3\textwidth}{\includegraphics[scale=0.3]{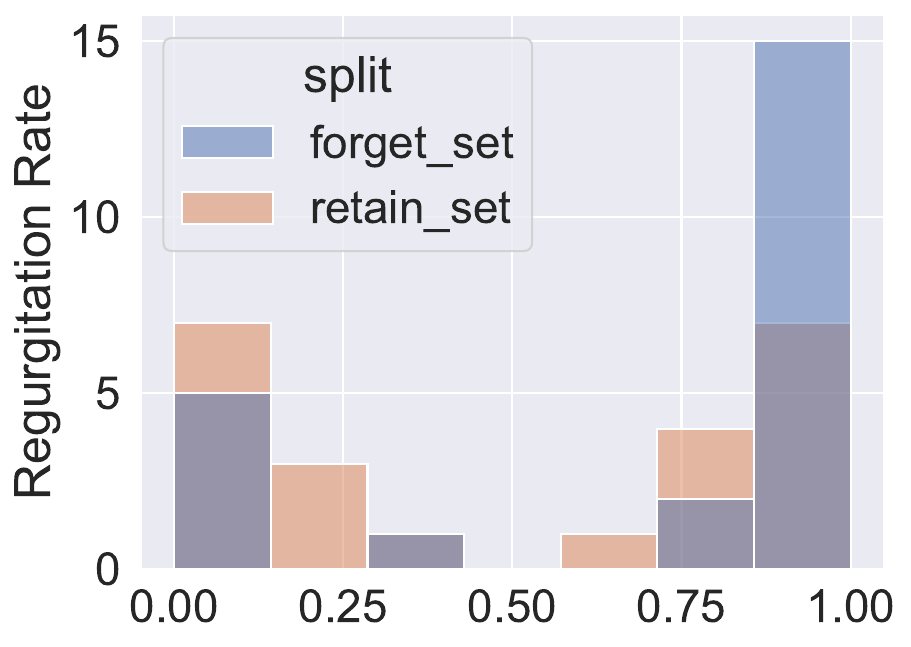}}\end{subfigure}
    \begin{subfigure}[b]{0.3\textwidth}{\includegraphics[scale=0.3]{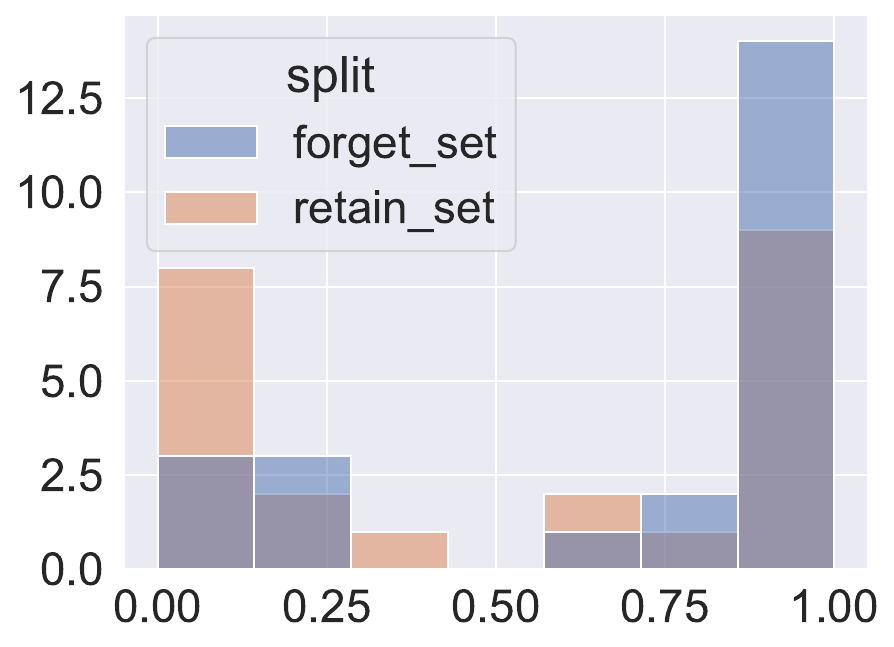}}\end{subfigure}
    \begin{subfigure}[b]{0.3\textwidth}{\includegraphics[scale=0.3]{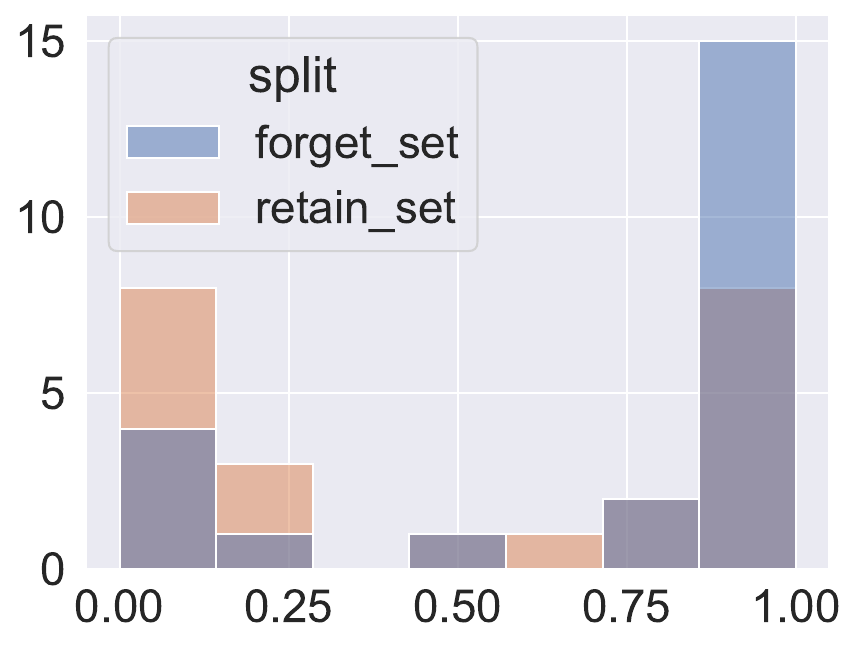}}\end{subfigure}
    
    \centering
    \begin{subfigure}[b]{0.3\textwidth}{\includegraphics[scale=0.3]{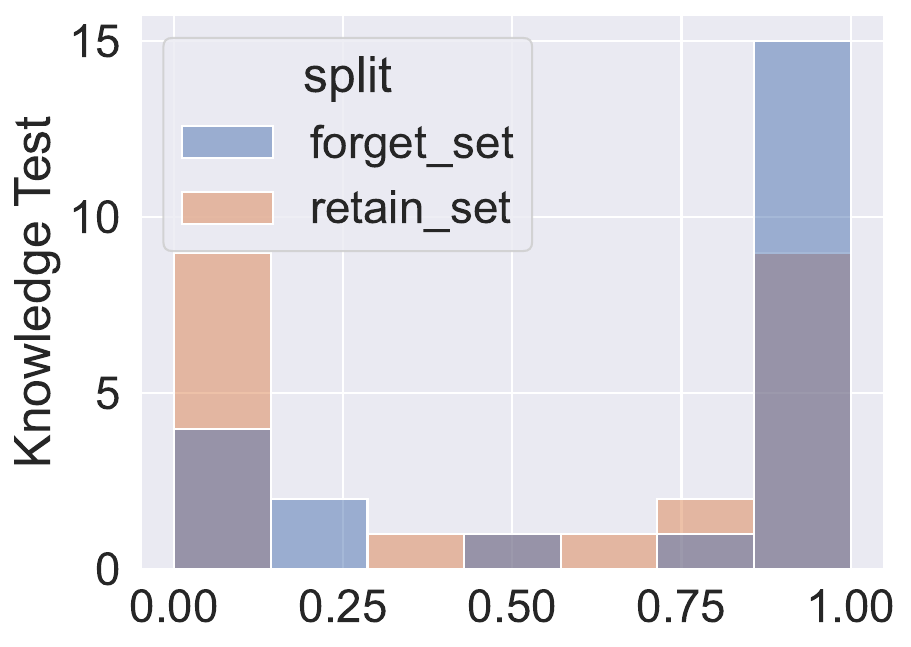}}\caption{Task1}\end{subfigure}
    \begin{subfigure}[b]{0.3\textwidth}{\includegraphics[scale=0.3]{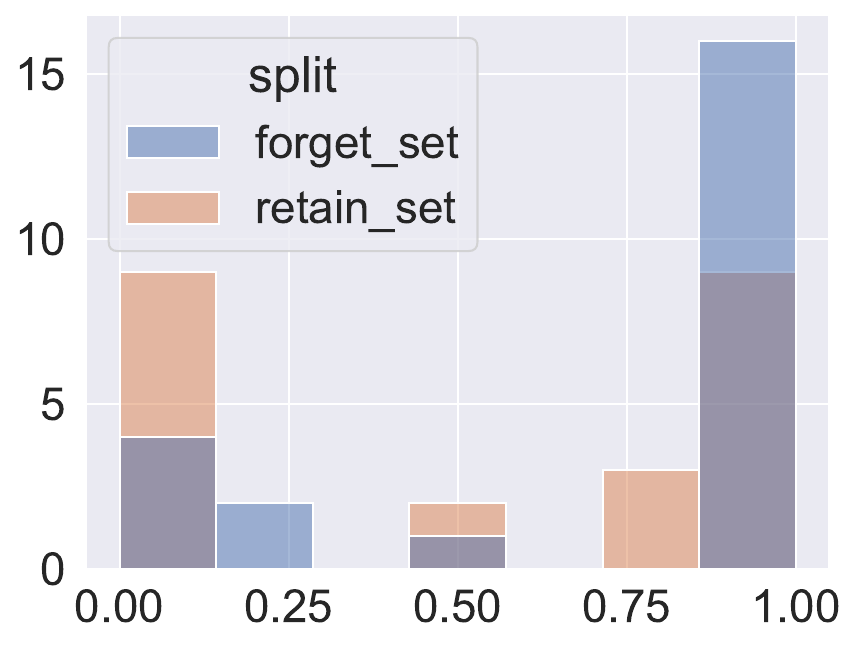}}\caption{Task2}\end{subfigure}
    \begin{subfigure}[b]{0.3\textwidth}{\includegraphics[scale=0.3]{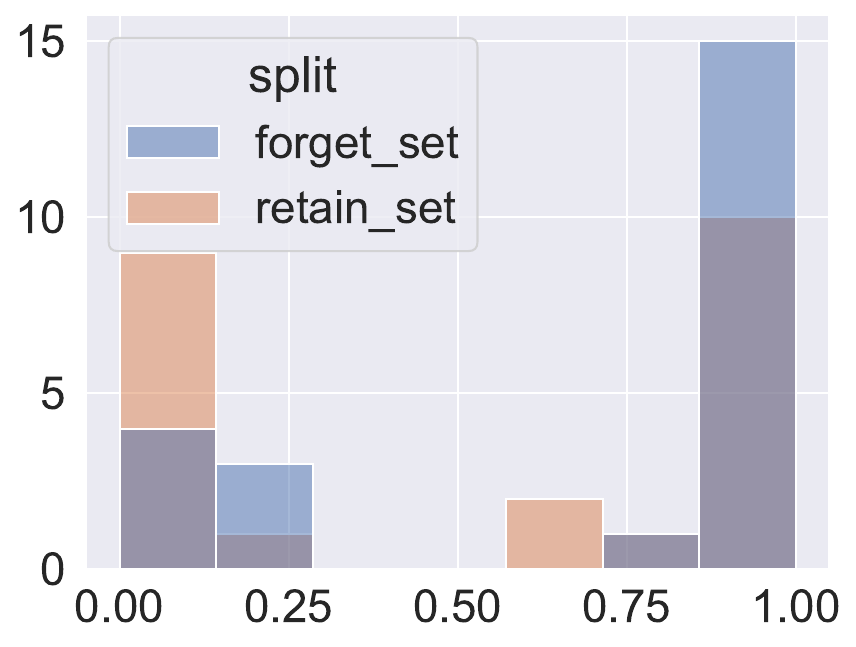}}\caption{Task3}\end{subfigure}
    \caption{Distribution of participant scores for forget and retain sets on the 1B model for all 6 sub-tasks. }
 \label{fig:1B_taskwise_histogram}
\end{figure*}

Finally, a handful of teams which were disqualified in 7B evals due to a drop in their MMLU utility recovered higher positions in the 1B leaderboard.
Notably, \textbf{SHA256} achieved a high Final Score (0.711), Task Aggregate (0.964), and MIA Score (0.894) with the 7B model. However, their MMLU score (0.275) dropped below the pre-defined threshold of 0.371, suggesting a substantial drop in overall model utility after unlearning. As a result, their system was regrettably disqualified in 7B evals but retained for 1B.

Table \ref{tab:subtask} presents task wise breakdown of top 5 teams in the 7B model. Results show that the top three systems achieve nearly perfect performance on the forget set, demonstrating the effectiveness of their methods in reducing regurgitation and removing knowledge from the LLMs. However, in several cases the performance on the retain sets drops considerably, suggesting \textit{over-unlearning}, leading to unintended forgetting of relevant information from the model. When comparing across tasks, Task 2 appears relatively easier than the other two tasks since it largely deals with short form, factual answers, with both \textbf{AILS-NTUA} and \textbf{YNU} achieving near-perfect scores in this task.


We plot histograms of team performances for both models in Figure \ref{fig:histograms}. 
Most teams score low on MIA, with only three teams scoring high on the 7B model while most others scored close to zero, suggesting imbalanced unlearning in these submissions. 
The MMLU scores for the 7B model are split into two clusters above and below the pre-defined threshold for rejection, with most submissions scoring above this threshold, suggesting delibrate parameter tuning to stop unlearning before this score drops below the threshold. For 1B model, since the base model performance on MMLU was already close to random chance, there is minimal impact due to the unlearning algorithms. 
The final score plots show an approximately bi-modal distribution, with a majority of teams with low scores except a select few which score highly.

We also plot distributions of sub-task wise performance for all teams for the two models in Figures \ref{fig:7B_taskwise_histogram} and \ref{fig:1B_taskwise_histogram}. We plot 1-test scores for the Forget set for easy comparison with the retain set. 
Across both models and in a majority of subtasks, the highest performing teams score considerably better with the forget set compared to retain set as observed in Table \ref{tab:subtask}. This is also due to over-unlearning in low scoring submissions which would remove the sensitive information but cause substantial degradations in the retain set as illustrated by a relatively uniform spread of retain set scores. 
We also observe an approximately bi-modal distribution across all tasks for the 1B model while for the 7B model some teams scored intermediate values. 


\section{Key Takeaways}

\paragraph{What were the key strategies explored by the teams?} The top team, along with a few others, applied gradient-based unlearning with low-rank adaptation (LoRA). These parameter-efficient updates enable the model to be fine-tuned efficiently, allowing for more iterations and the use of a larger retain dataset. Similarly, several teams developed selective unlearning techniques to identify and target specific parameters or layers for unlearning. Finally, balancing between over or under unlearning is critical and several teams fail to address it, causing low MMLU or MIA scores respectively.

\paragraph{Is the task solved?} While the top-performing team achieved high scores, its utility (measured by MMLU) still experienced a notable drop, from 0.494 to 0.443. Their model checkpoint was also reported to generate garbage tokens with specific prompts, suggesting some degree of model degradation due to unlearning. In contrast, other teams maintained utility but did not improve on MIA or task aggregate scores. This highlights that balancing utility and unlearning effectiveness remains a challenging and open task for future work.


\paragraph{What can we do differently?}
Several participants reported not having access to a multi-gpu training environment, and submitted code which was not tested with Deepspeed. As a result, substantial manual effort was invested in modifying almost all submitted code files to train on our evaluation environment. This can be avoided by using suitable platforms such as Huggingface competitions, which will be explored in future cycles. 

\section{Conclusion and Future Improvements}
This paper summarizes SemEval-2025 Task 4 on unlearning sensitive content from LLMs. Our task presents a significant challenge, as most systems struggle to maintain model utility while completely unlearning unwanted information. However, we received several innovative solutions which made strong contributions towards solving this task. We hope our challenge and the associated benchmark inspire further research into efficient methods for unlearning sensitive content from LLMs. 

We note several avenues for future exploration:
\begin{compactenum}
\item \textbf{Evaluation metrics.} Outside LLMs, unlearning literature typically uses some form of statistical hypothesis testing between the model posteriors from the unlearned and the retrained (i.e. trained without the sensitive information) model candidates. However, this is not feasible for LLMs since the model would have to be trained from ground up, including pretraining which is a computationally expensive undertaking and not explored in current LLM unlearning research. 
\item \textbf{Larger model checkpoints:} We limited our challenge to 7 and 1 billion parameter models due to limited compute availability with most participants. In future work we may expand on this challenge by inviting a subset of teams to onboard to specialized compute platforms to motivate further research on unlearning larger models.
\item \textbf{Unlearning other attributes:} Unlearning of sensitive information or a class of model capabilities (such as coding in a specific language). 

\end{compactenum}
\bibliography{custom}


\begin{figure*}[!ht]
    \centering
    \includegraphics[width=\linewidth]{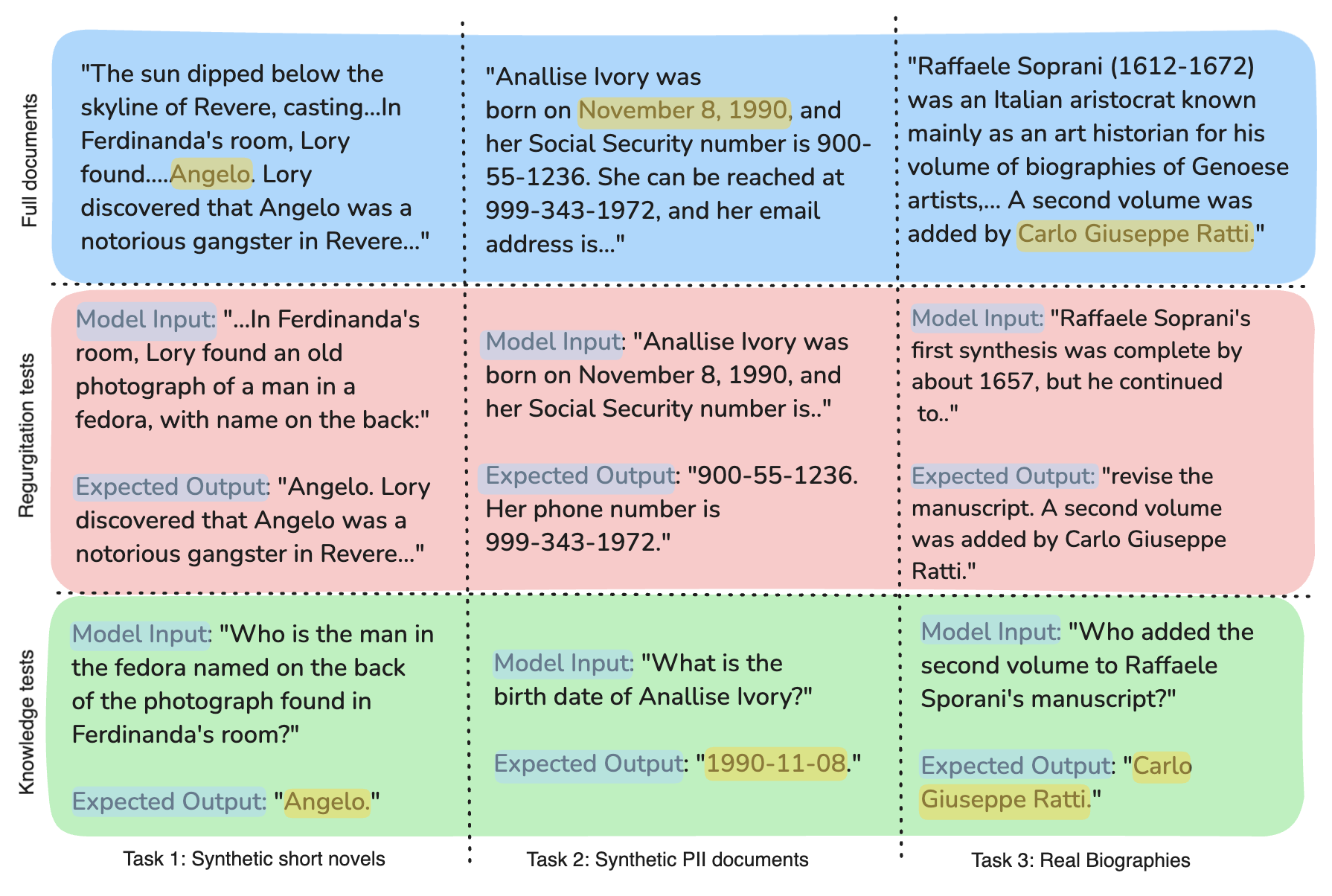}
    \caption{Examples of full documents and test prompts for the three tasks covered in this challenge. The figure is quoted from \cite{ramakrishna2025lumellmunlearningmultitask}.}
    \label{fig:lume_approach}
\end{figure*}

\appendix

\section{Examples}

Figure \ref{fig:lume_approach} shows examples from all six subtasks in our benchmark. 

\section{Creative Content Generation Prompt}
\label{sec:long_form_section}
\begin{minipage}{\linewidth}
\begin{lstlisting}
Model Input:
Create a short novel of at least 150 words. The novel should be from fantasy genre and set in the city of Atlantis. It should include following characters: Kyle, Stan, Kenny and Eric.
    
Model Output: 
A short story about four friends lost in the magical realm below the oceans, known to ousiders as Atlantis. Kyle had always held a deep fascination for the deep blue ocean, and this naturally led him to take up a major in oceanic studies...
\end{lstlisting}
\end{minipage}

\section{Personal Biography Generation Prompt}
\label{sec:short_form_section}
\begin{minipage}{\linewidth}
\begin{lstlisting}
Model Input:
Create a biography for Jon Smith with date of birth: 1/2/1989, SSN: 900123456, phone number: 0987654321, email: jon_smith@me.com, home address: 10 Summertime Lane, New York City, NY, USA. 
    
Model Output: 
Jon Smith was both in New York City on the first of February in 1989,...
\end{lstlisting}
\end{minipage}

\section{Question Generation Prompt}
\label{sec:question_generation_prompt}
\begin{minipage}{\linewidth}
\begin{lstlisting}
Model Input:
You are given a short story. First, find all the proper nouns in this story. If it does not contain a proper noun, say "I can't use this statement since it does not contain any proper nouns.". If it contains proper nouns, use your reasoning to create an unambiguous question, for which there would be *only* one answer. Give a concise answer (i.e. one word or phrase) which accurately answers the question. If you cannot create such an unambiguous question, say "I'm unable to create an unambiguous question for this story". Use the examples below for reference.

Examples:
1. Example #1
2. Example #2
3. Example #3
4. Example #4
5. Example #5

Here's the story: <input_story>. Generate a question with an unambiguous answer using this story. 
\end{lstlisting}
\end{minipage}


\section{Task Wise Benchmark Results}
\label{sec:task_wise_benchmark_results}

Figures \ref{fig:7B_taskwise_histogram} and \ref{fig:1B_taskwise_histogram} show task wise distributions on forget and retain sets for all benchmarked unlearning algorithms. 

\begin{figure*}
    \centering
    \begin{subfigure}[b]{0.22\textwidth}{\includegraphics[scale=0.25]{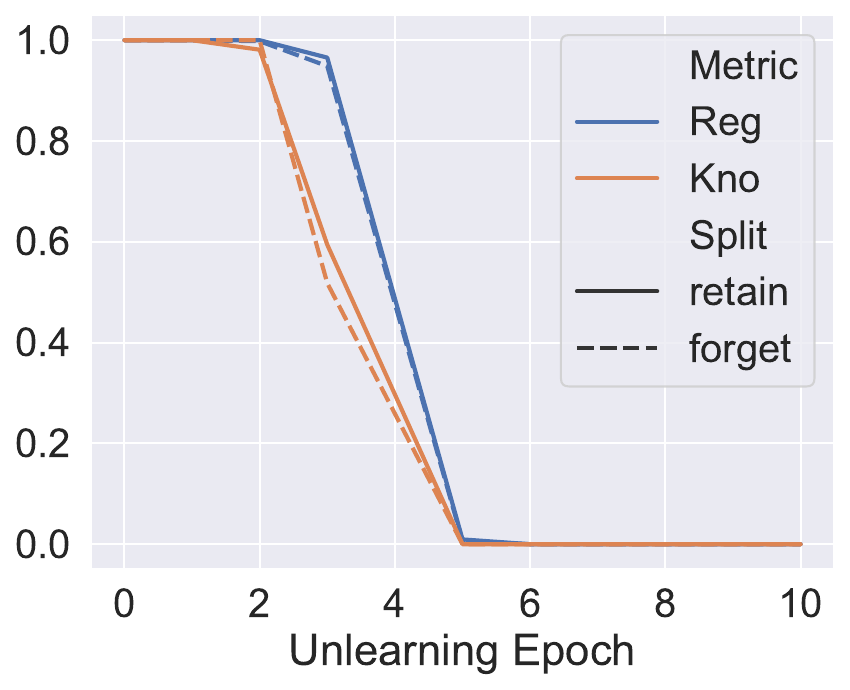}}\end{subfigure}
    \begin{subfigure}[b]{0.22\textwidth}{\includegraphics[scale=0.25]{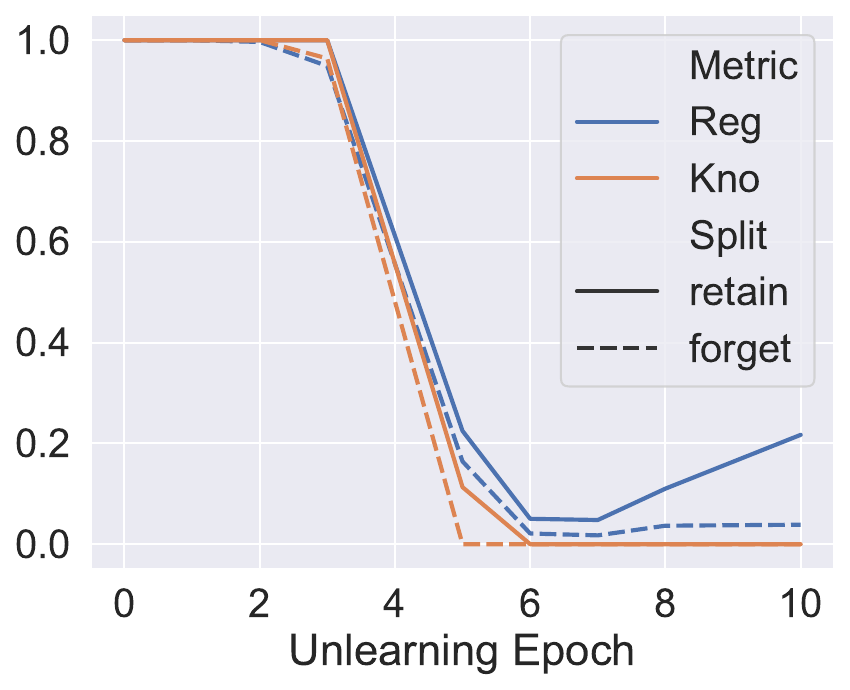}}\end{subfigure}
     \begin{subfigure}[b]{0.22\textwidth}{\includegraphics[scale=0.25]{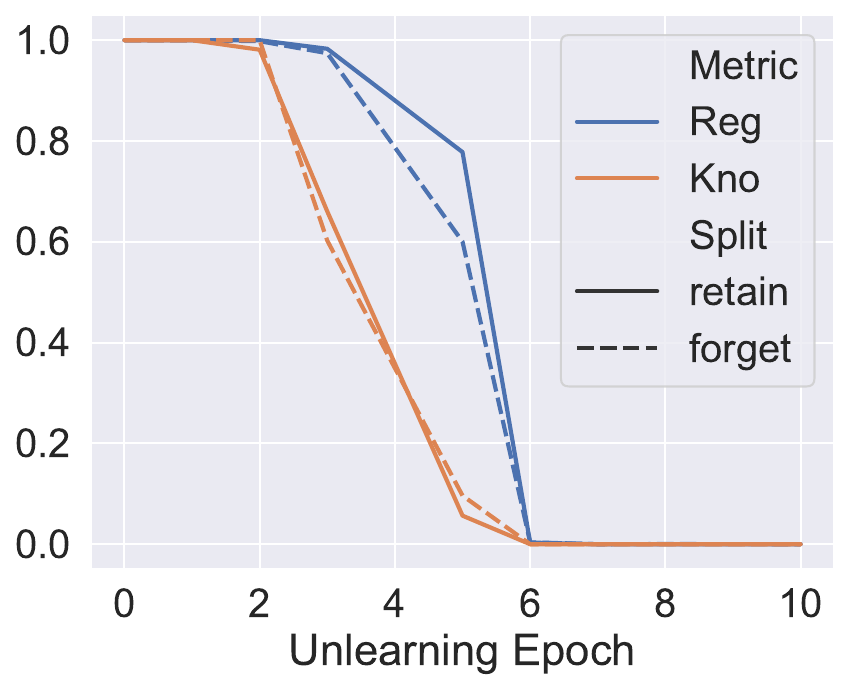}}\end{subfigure}
     \begin{subfigure}[b]{0.22\textwidth}{\includegraphics[scale=0.25]{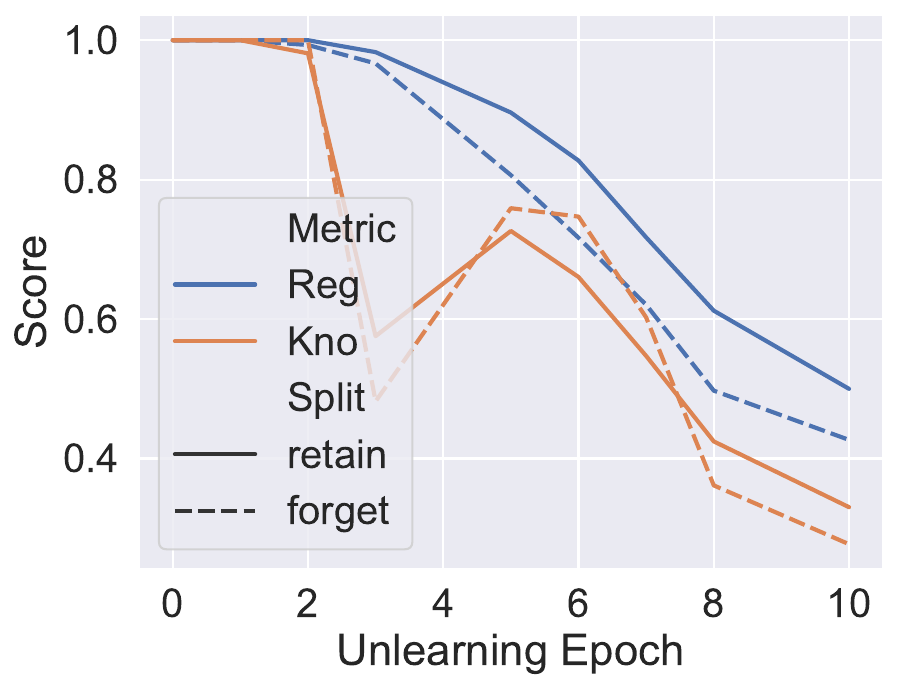}}\end{subfigure}

     \begin{subfigure}[b]{0.22\textwidth}{\includegraphics[scale=0.25]{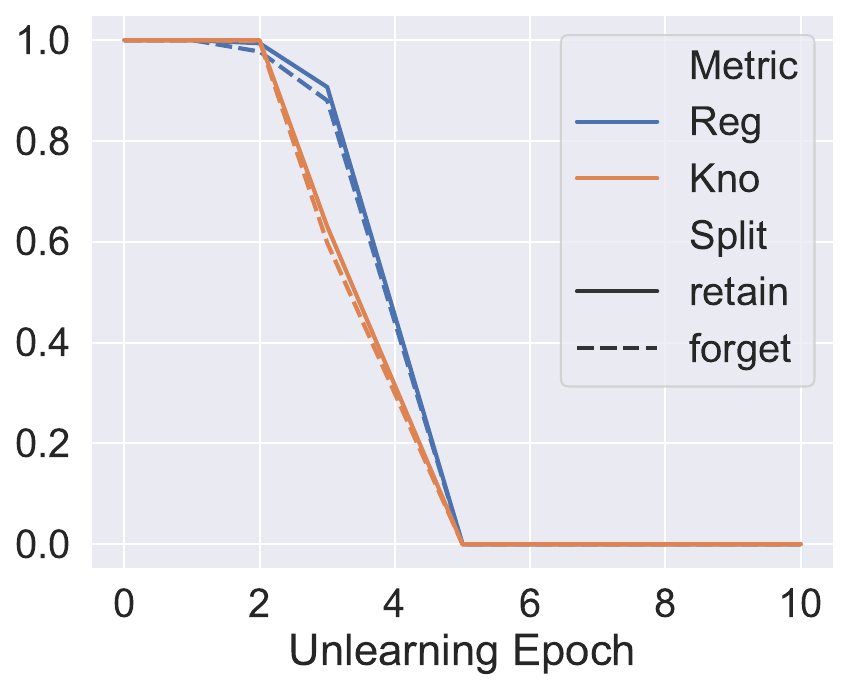}}\end{subfigure}
    \begin{subfigure}[b]{0.22\textwidth}{\includegraphics[scale=0.25]{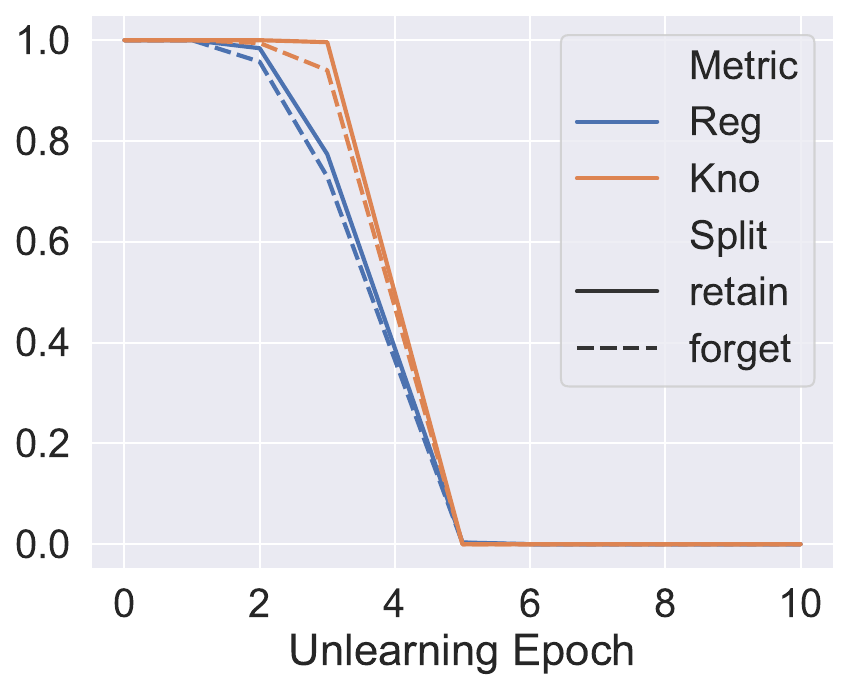}}\end{subfigure}
     \begin{subfigure}[b]{0.22\textwidth}{\includegraphics[scale=0.25]{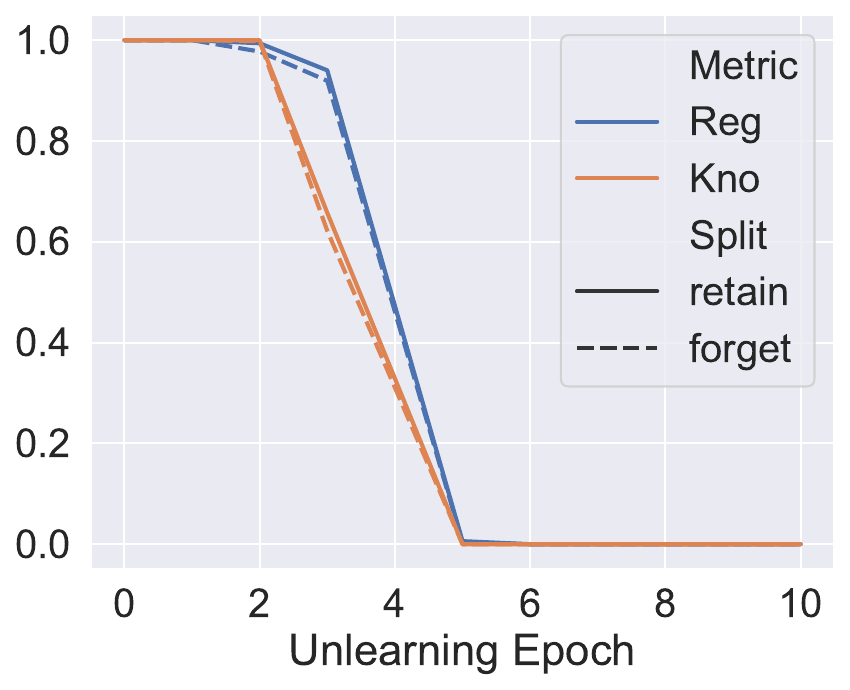}}\end{subfigure}
     \begin{subfigure}[b]{0.22\textwidth}{\includegraphics[scale=0.25]{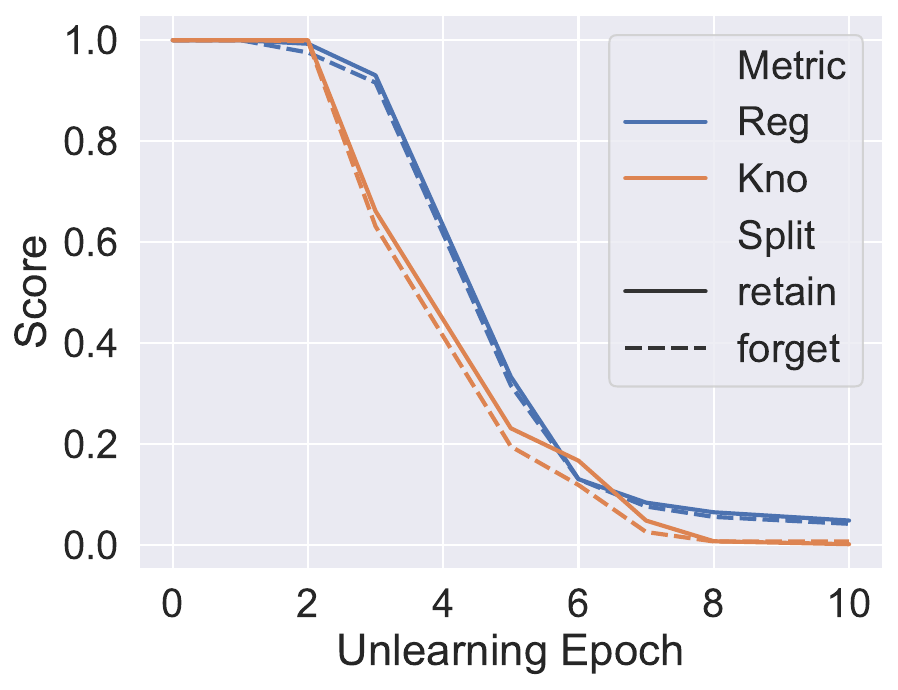}}\end{subfigure}

     \begin{subfigure}[b]{0.22\textwidth}{\includegraphics[scale=0.25]{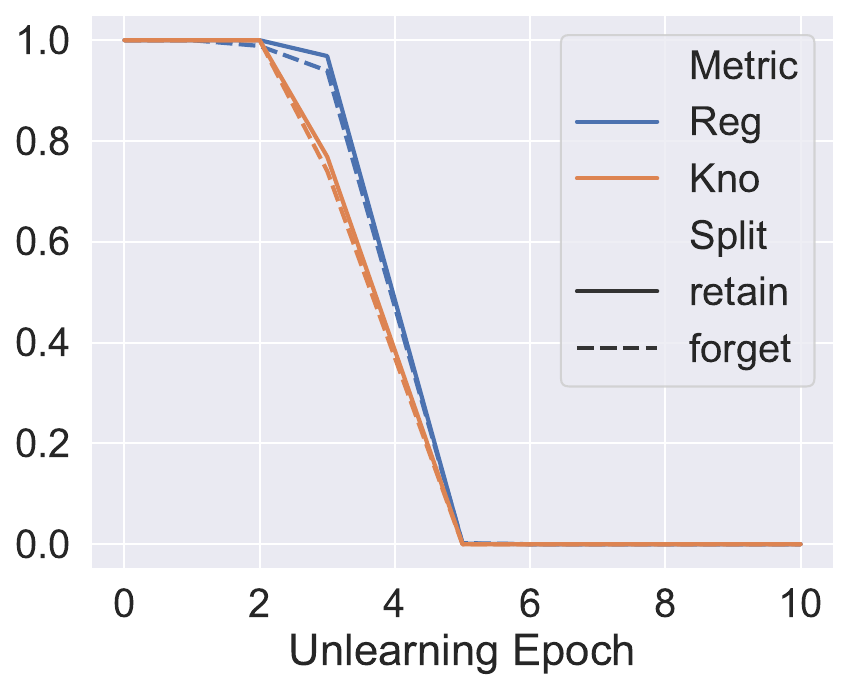}}
     \caption{GA}\end{subfigure}
    \begin{subfigure}[b]{0.22\textwidth}{\includegraphics[scale=0.25]{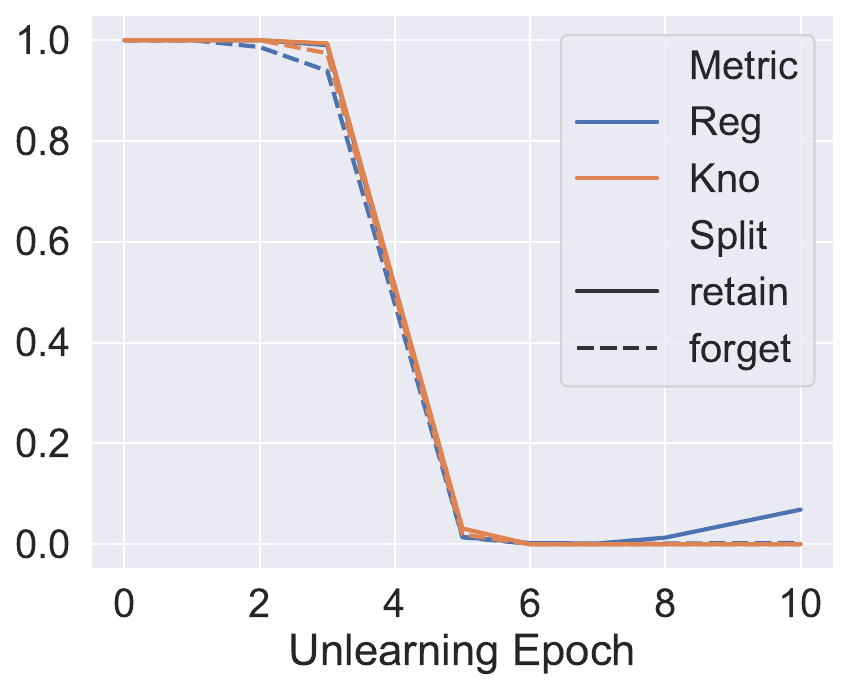}}
     \caption{GD}\end{subfigure}
     \begin{subfigure}[b]{0.22\textwidth}{\includegraphics[scale=0.25]{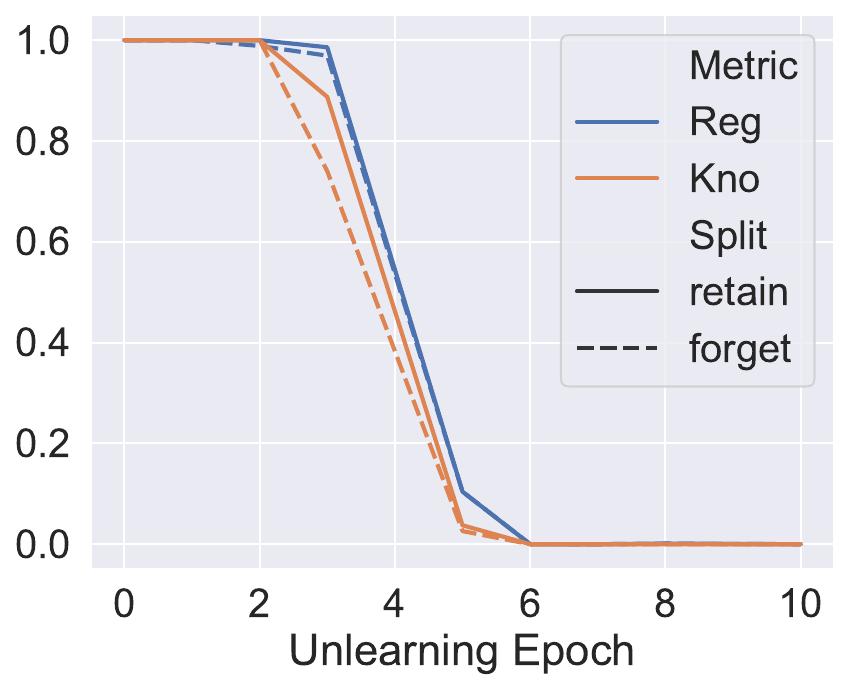}}
     \caption{KL}\end{subfigure}
     \begin{subfigure}[b]{0.22\textwidth}{\includegraphics[scale=0.25]{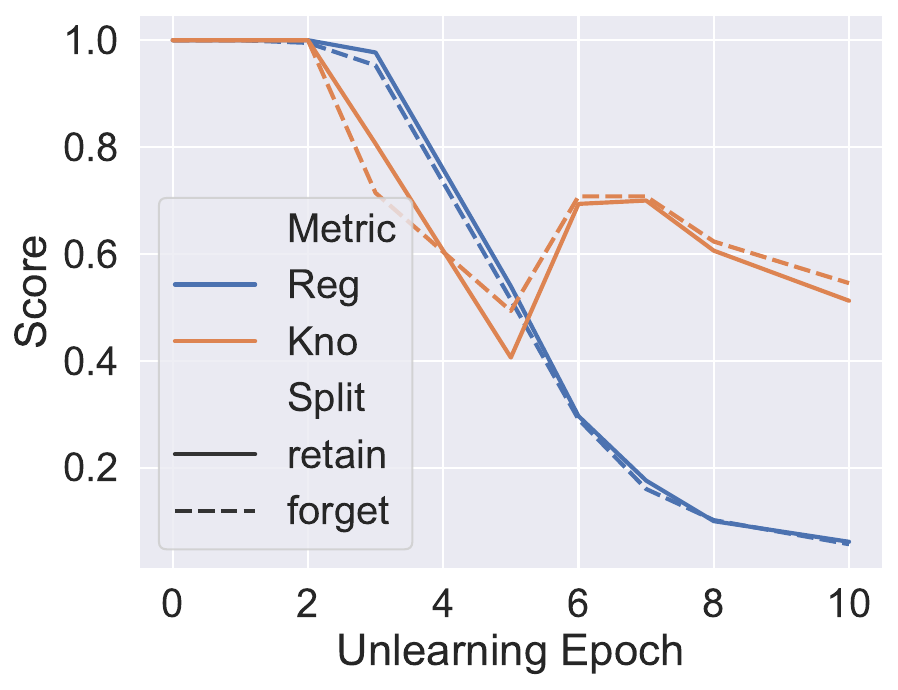}}
     \caption{NPO}\end{subfigure}
    \caption{Performance on \textit{retain} and \textit{forget} subsets for 7B model for benchmarked unlearning algorithms for Tasks 1 to 3 (respectively from top to bottom). Reg: Regurgitation Rate ($r$), Kno: Knowledge Accuracy ($t$). Split refers to data subset (forget or retain) used in evaluations.}
    \label{fig:7Bresults}
\end{figure*}

\begin{figure*}
    \centering
    \begin{subfigure}[b]{0.22\textwidth}{\includegraphics[scale=0.25]{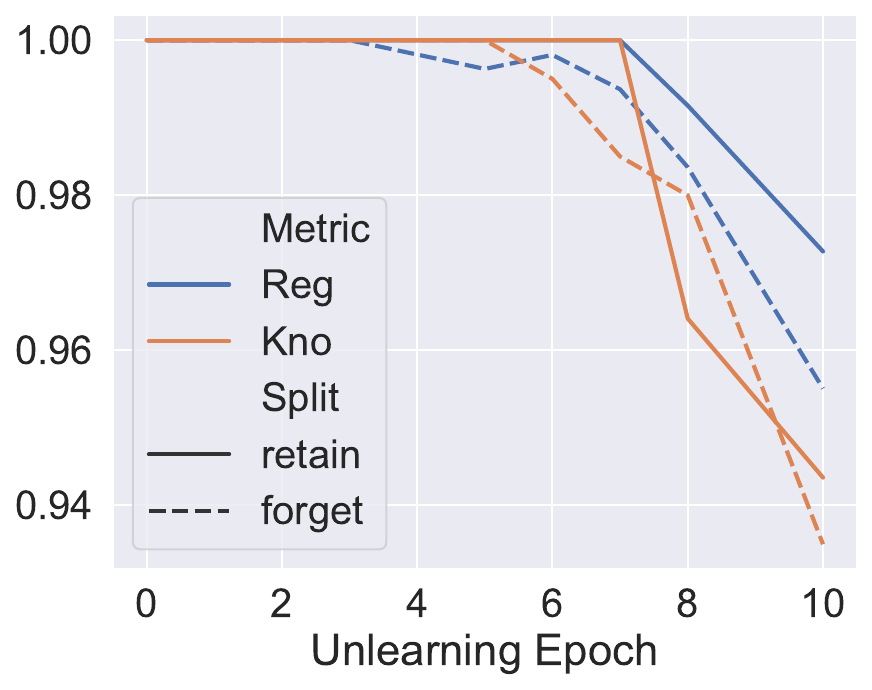}}\end{subfigure}
    \begin{subfigure}[b]{0.22\textwidth}{\includegraphics[scale=0.25]{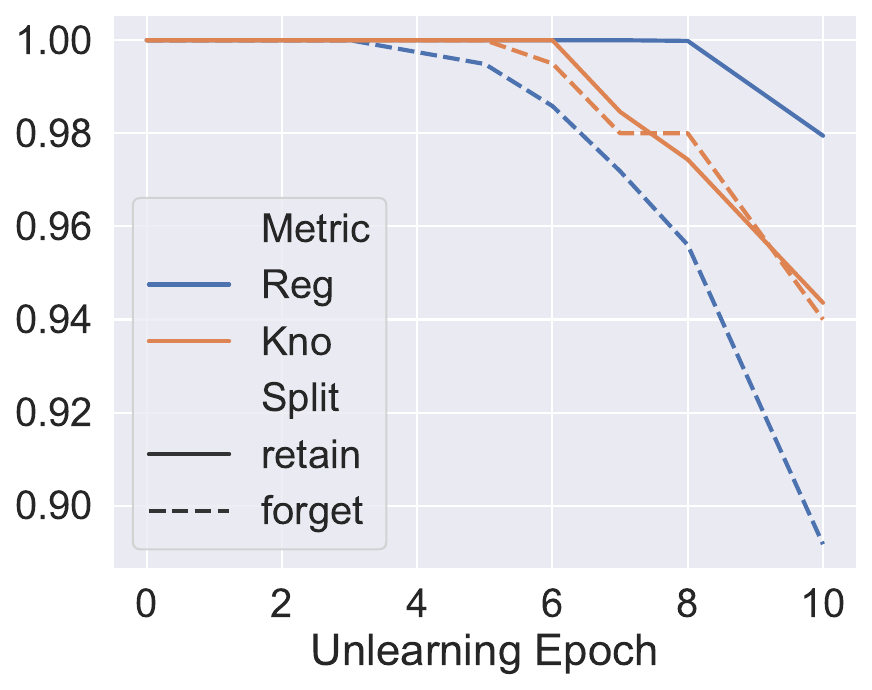}}\end{subfigure}
     \begin{subfigure}[b]{0.22\textwidth}{\includegraphics[scale=0.25]{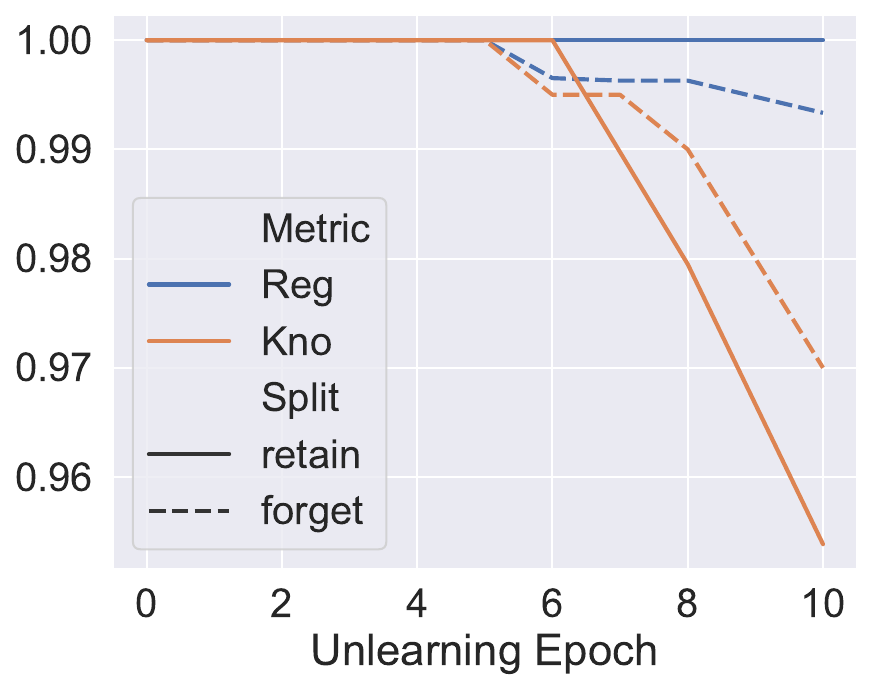}}\end{subfigure}
     \begin{subfigure}[b]{0.22\textwidth}{\includegraphics[scale=0.25]{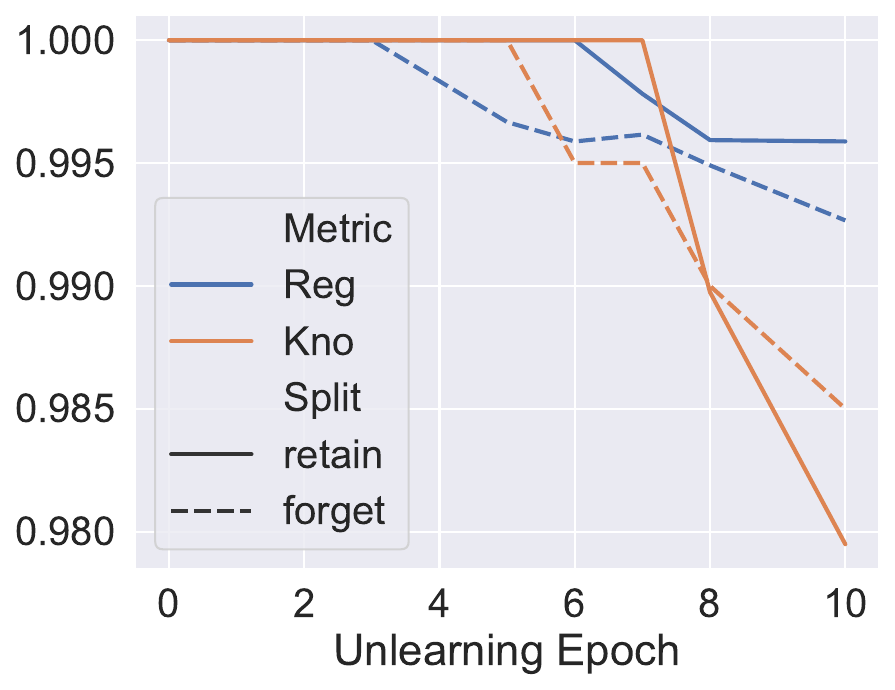}}\end{subfigure}

     \begin{subfigure}[b]{0.22\textwidth}{\includegraphics[scale=0.25]{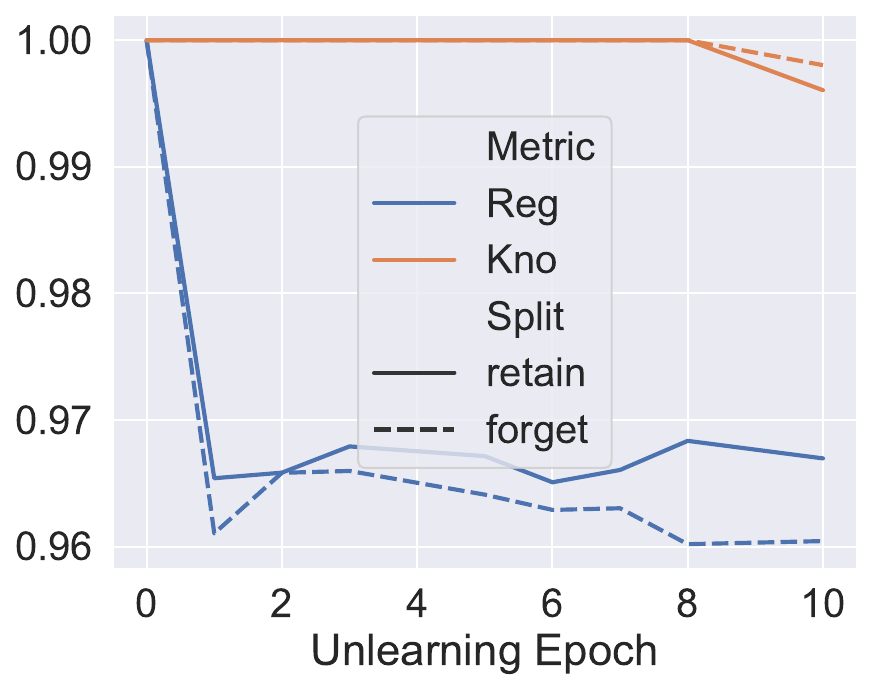}}\end{subfigure}
    \begin{subfigure}[b]{0.22\textwidth}{\includegraphics[scale=0.25]{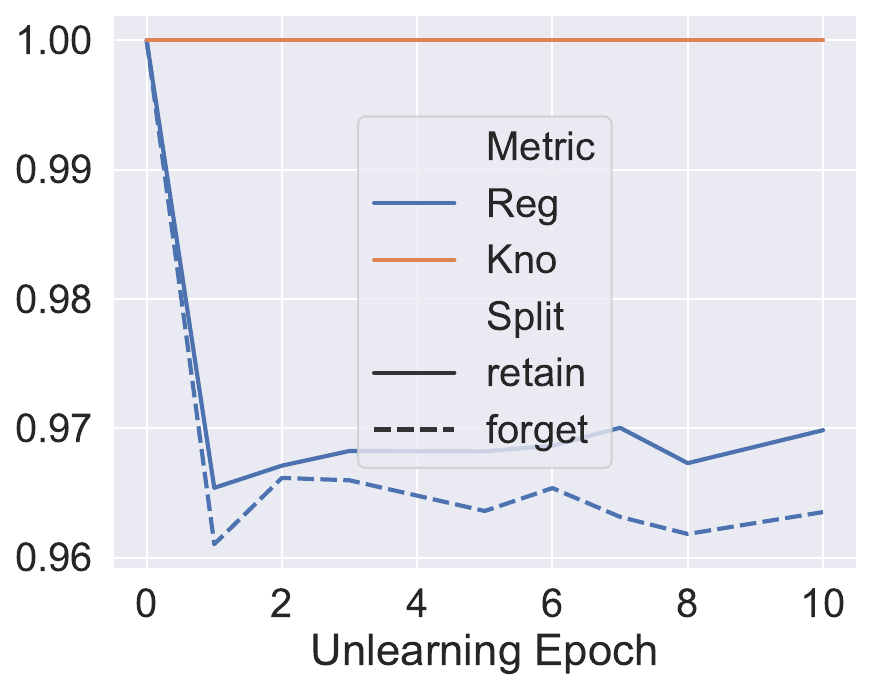}}\end{subfigure}
     \begin{subfigure}[b]{0.22\textwidth}{\includegraphics[scale=0.25]{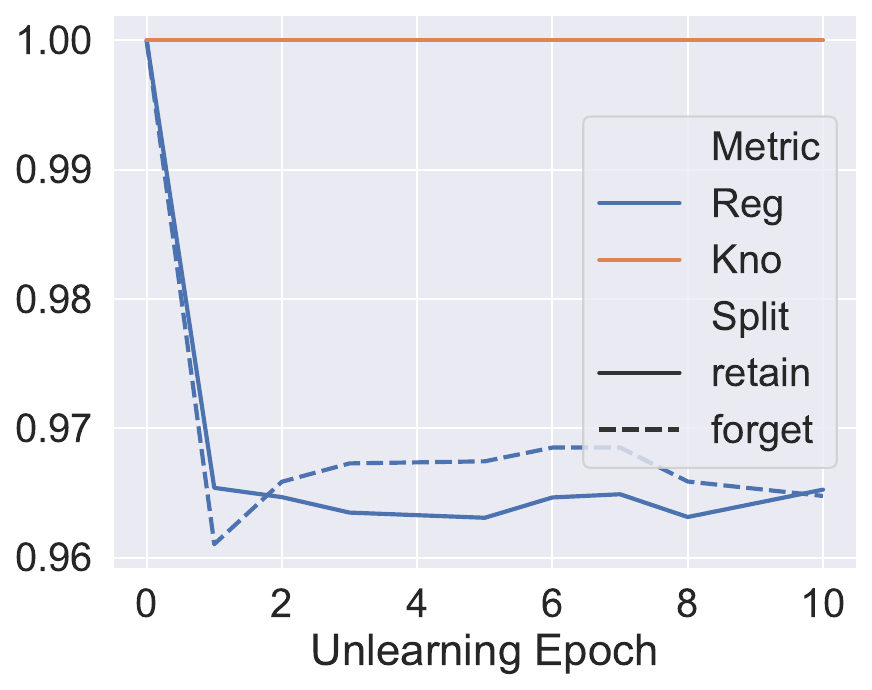}}\end{subfigure}
     \begin{subfigure}[b]{0.22\textwidth}{\includegraphics[scale=0.25]{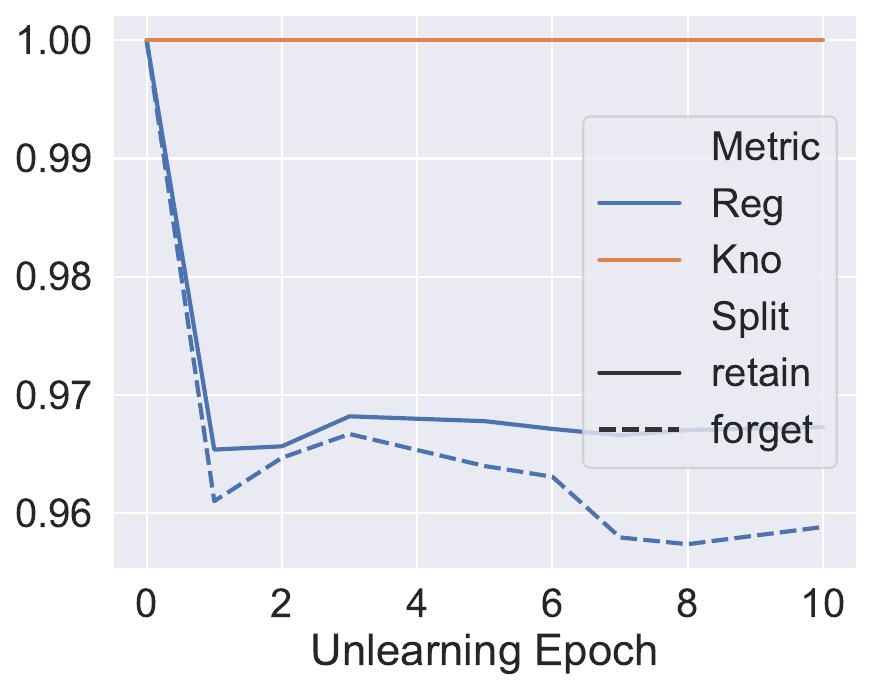}}\end{subfigure}

     \begin{subfigure}[b]{0.22\textwidth}{\includegraphics[scale=0.25]{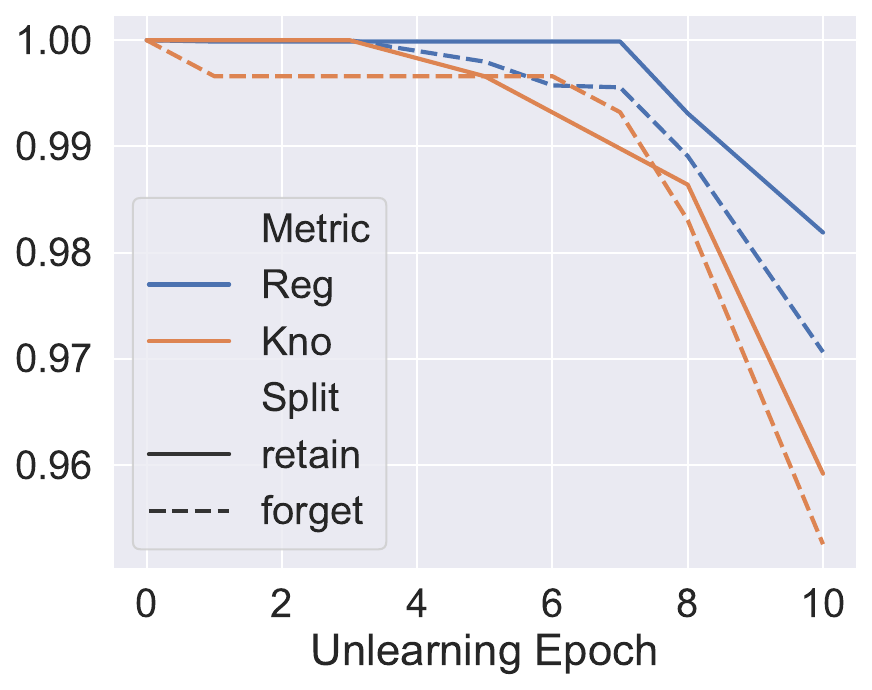}}
     \caption{GA}\end{subfigure}
    \begin{subfigure}[b]{0.22\textwidth}{\includegraphics[scale=0.25]{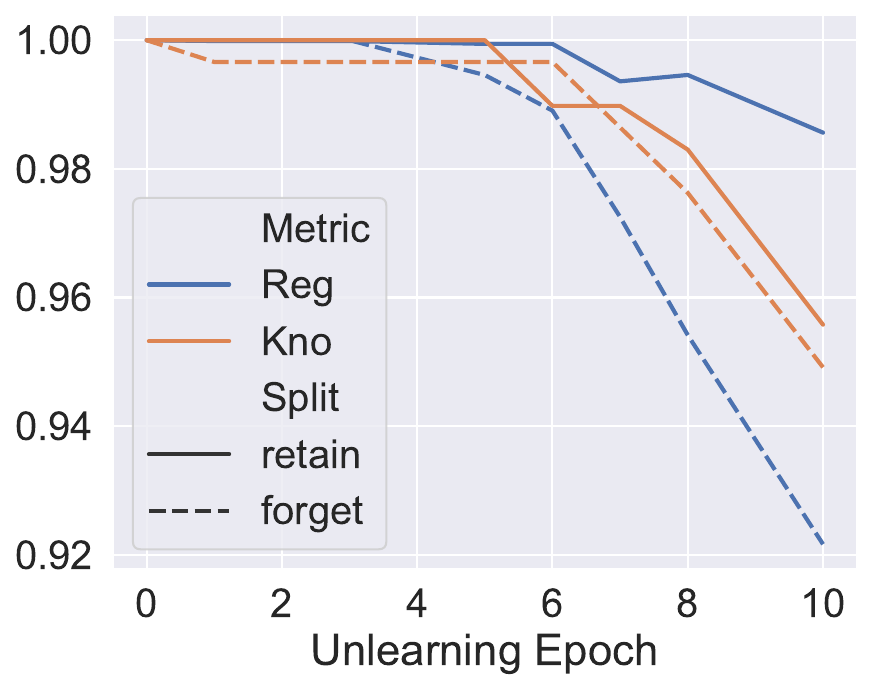}}
     \caption{GD}\end{subfigure}
     \begin{subfigure}[b]{0.22\textwidth}{\includegraphics[scale=0.25]{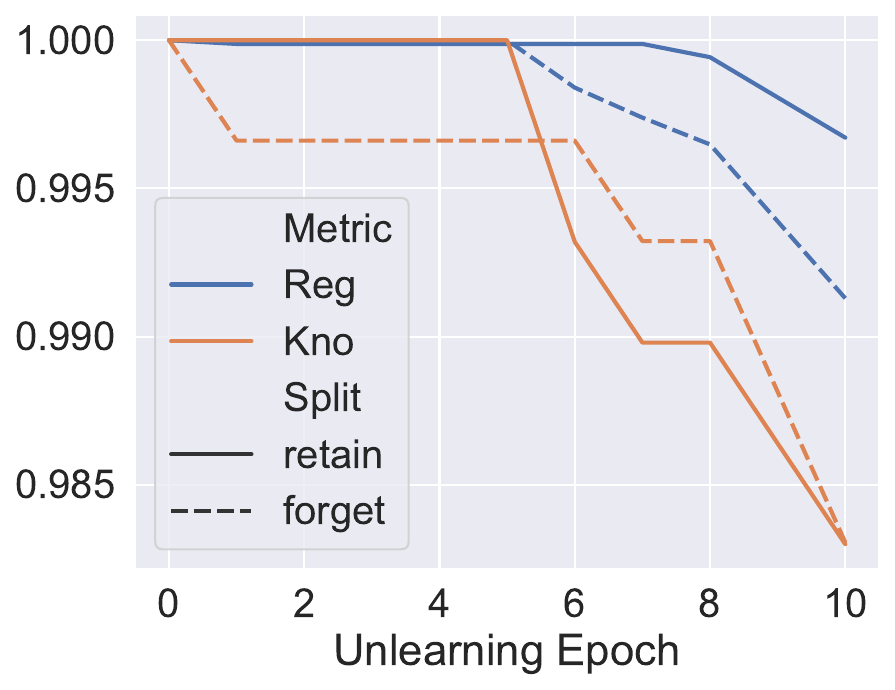}}
     \caption{KL}\end{subfigure}
     \begin{subfigure}[b]{0.22\textwidth}{\includegraphics[scale=0.25]{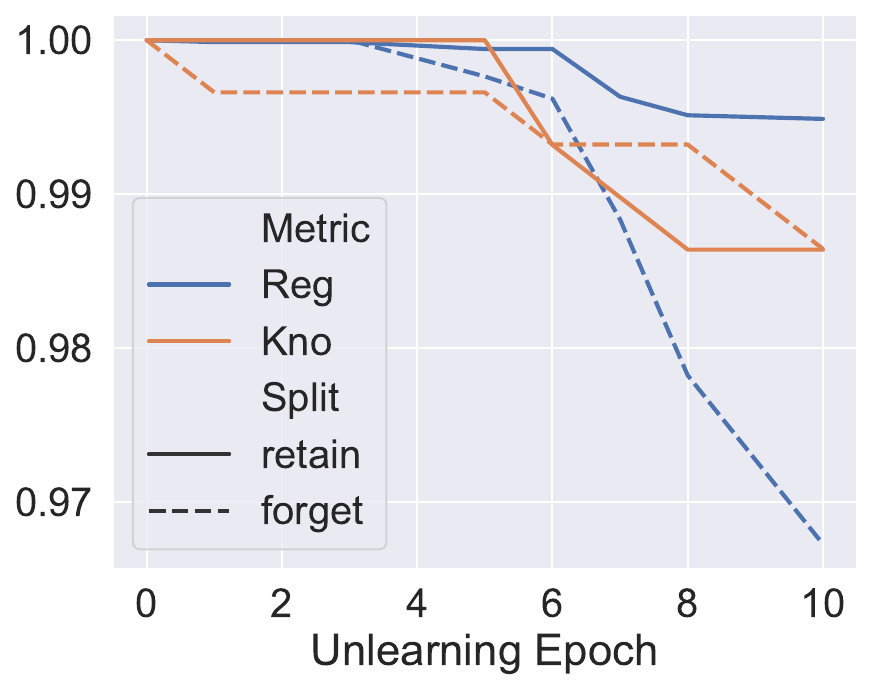}}
     \caption{NPO}\end{subfigure}
    \caption{Performance on \textit{retain} and \textit{forget} subsets for 1B model for benchmarked unlearning algorithms for Tasks 1 to 3 (respectively from top to bottom). Reg: Regurgitation Rate ($r$), Kno: Knowledge Accuracy ($t$). Split refers to data subset (forget or retain) used in evaluations.}
    \label{fig:1Bresults}
\end{figure*}

\section{System descriptions}

\begin{table*}[h!]
\centering
\begin{tabular}{m{5cm} m{10cm}} 
\hline
\toprule
\multicolumn{1}{c}{\textit{Team}}       & \textit{Core strategy} \\ \midrule
AILS-NTUA                       & Iterative unlearning on carefully sampled chunks of forget set, mixed with a larger volume of retain set                              \\
ZJUKLAB                         & Two distinct NPO+KL+GD trained models are merged to balance under/over-unlearning between them.                                          \\
YNU            & Unlearning with random tokens followed by alternating GA/GD on forget/retain samples.                                                    \\
Mr. Snuffleupagus               & Adaptive RMU on three layers selected using validation set.                                                                              \\
ishumei-Chinchunmei             & Alternate formulation for unlearning loss as reciprocal of gradient descent (instead of inverted sign as is done in GA).                 \\
GUIR             & Unlearning with adaptive tuning of weights for forget and retain sets                                                                    \\
GIL-IIMAS UNAM                  & Selective GA followed by GD (7B) and Task vector from forget set subtracted for unlearning (1B)                                          \\
Atyaephyra                      & NPO using LoRA adapters (for compute efficiency), with reference probability obtained by removing LoRA adapters (for memory efficiency). \\
Lacuna Inc.                     & Selective parameter unlearning on parameters not relevant for retain set, selected using Fisher Information Matrix                       \\
NLPART                          & NPO+SFT on deflection strings.                                                                                                           \\
JU-CSE-NLP'25                   & Normalized Gradient Difference with AutoLR \cite{bu2024unlearningmultitaskoptimizationnormalized}                                                                  \\
SHA256                          & Causal mediation to identify first 5 layers as most impactful, followed by unlearning using GD on these layers.                          \\
NeuroReset                      & GA on forget set followed by GD on retain set (3 epochs each)                                                                            \\
Cyber for AI                    & Gradient Difference followed by gradient ascent.                                                                                         \\
MALTO                           & Distillation from aggregated probability from incompetent (forget set) and competent (retain set) teachers.                              \\
NEKO & GA with KL regularization on retain set from reference model.                                                                            \\
DUTir                           & Selective parameter unlearning on parameters identified using gradients for forget and retain sets.                                      \\
AI4PC            & Distillation from two models enhanced on forget and retain sets separately.                                \\ \bottomrule                             
\end{tabular}
\caption{Brief summaries of key strategys employed by all participating teams.}
\label{tab:strategy}
\end{table*}

We provide brief descriptions for submissions from all participants in Table \ref{tab:strategy}.

\end{document}